\documentclass{article}

\usepackage[preprint]{tackling_climate_workshop_style}

\usepackage{hyperref}
\usepackage[utf8]{inputenc} % allow utf-8 input
\usepackage[T1]{fontenc}    % use 8-bit T1 fonts
\usepackage{xurl}
\usepackage{booktabs}       % professional-quality tables
\usepackage{amsfonts}       % blackboard math symbols
\usepackage{nicefrac}       % compact symbols for 1/2, etc.
\usepackage{microtype}      % microtypography.
\usepackage{graphicx}       % figures
\usepackage{makecell}
\usepackage{multirow}
\usepackage{caption}
\usepackage{subcaption}
\usepackage{array}
\usepackage{tikz}

\title{Satellite Sunroof: High-res Digital Surface Models and Roof Segmentation for Global Solar Mapping}

\author{%
Vishal Batchu*$^{1}$ \quad Alex Wilson*$^{1}$ \quad Betty Peng$^1$ \quad Carl Elkin$^1$ \quad Umangi Jain$^2$ \\
\textbf{Christopher Van Arsdale}$^1$ \quad \textbf{Ross Goroshin}$^1$ \quad \textbf{Varun Gulshan}$^1$ \\
$^1$Google \quad $^2$University of Toronto$^\dagger$ \\
\texttt{\{vishalbatchu,alexwilson,bettypeng,celkin\}@google.com}\\
\texttt{\{umangi.jain\}@mail.utoronto.ca}\\
\texttt{\{cvanarsdale,goroshin,varungulshan\}@google.com}\\
$*$ Equal contribution from these authors.\\
$^\dagger$ Work done while at Google.
}

\begin{document}

\maketitle

\begin{abstract}
  The transition to renewable energy, particularly solar, is key to mitigating climate change. Google's Solar API aids this transition by estimating solar potential from aerial imagery, but its impact is constrained by geographical coverage. This paper proposes expanding the API's reach using satellite imagery, enabling global solar potential assessment. We tackle challenges involved in building a Digital Surface Model (DSM) and roof instance segmentation from lower resolution and single oblique views using deep learning models. Our models, trained on aligned satellite and aerial datasets, produce 25cm DSMs and roof segments. With \textasciitilde 1m DSM MAE on buildings, \textasciitilde 5$^\circ$ roof pitch error and \textasciitilde 56\% IOU on roof segmentation, they significantly enhance the Solar API's potential to promote solar adoption.
\end{abstract}

\section{Introduction}

Google Maps Platform Solar API \citep{GoogleSolarAPI} aims to increase the speed and depth of rooftop solar photo-voltaic roll-out by accurately estimating the solar potential of all suitable buildings worldwide using high quality aerial imagery. Since its release, we estimate that Solar API has been used in over 1 million residential solar projects in US, Europe and Japan. Previous work \citep{goroshin2023estimating} demonstrated the potential of such mapping with lower quality aerial imagery in the US and parts of Europe increasing data coverage by 10x over high quality aerial coverage. With satellite imagery, Solar API has the potential to increase its coverage by a further 1B buildings, focusing on the global south (20+ countries across South America, Asia, Africa, Australia and Europe). This expansion represents a further 10x increase in potential area coverage over high quality aerial imagery and enables a higher refresh rate to assess changes in solar potential over time.

Accurate rooftop geometry and shading analysis are crucial for the Solar API pipeline, which relies on precise Digital Surface Models (DSMs) and roof segmentation. We propose training deep learning models to predict DSMs and roof segments from satellite imagery, using high-quality aerial imagery as labels. Our method addresses challenges inherent in satellite data, such as lower resolution, oblique angles, and temporal discrepancies compared to aerial labels. Building on prior work \citep{goroshin2023estimating}, we demonstrate the potential of Solar API to enhance global solar potential assessment, advancing the transition to sustainable energy.

\begin{figure}[h!]
    \centering
    \hspace*{-0.03\textwidth}
    \begin{tabular}{>{\centering\arraybackslash}m{0.29\textwidth} >{\centering\arraybackslash}m{0.003\textwidth} >{\centering\arraybackslash}m{0.29\textwidth} >{\centering\arraybackslash}m{0.003\textwidth} >{\centering\arraybackslash}m{0.29\textwidth}}

        \begin{tabular}{c}
            \subcaptionbox{Off-nadir satellite RGB\label{fig:subfig1}}{\includegraphics[width=0.29\textwidth]{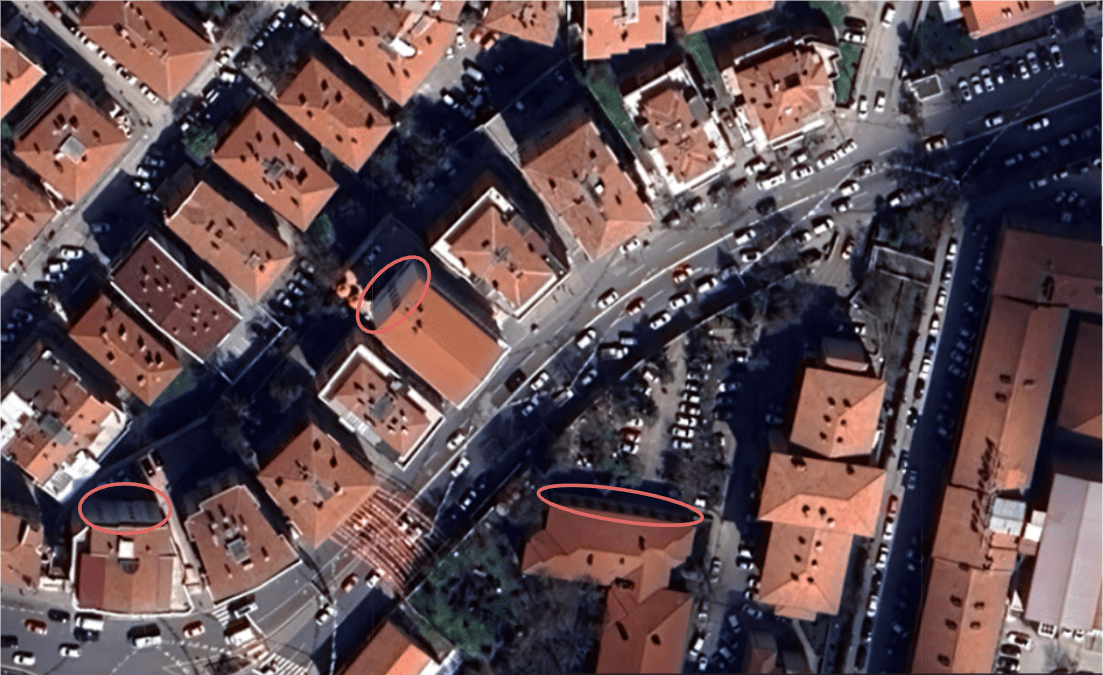}} \\
            + \\
            \subcaptionbox{(Optional) satellite DSM\label{fig:subfig2}}{\includegraphics[width=0.29\textwidth]{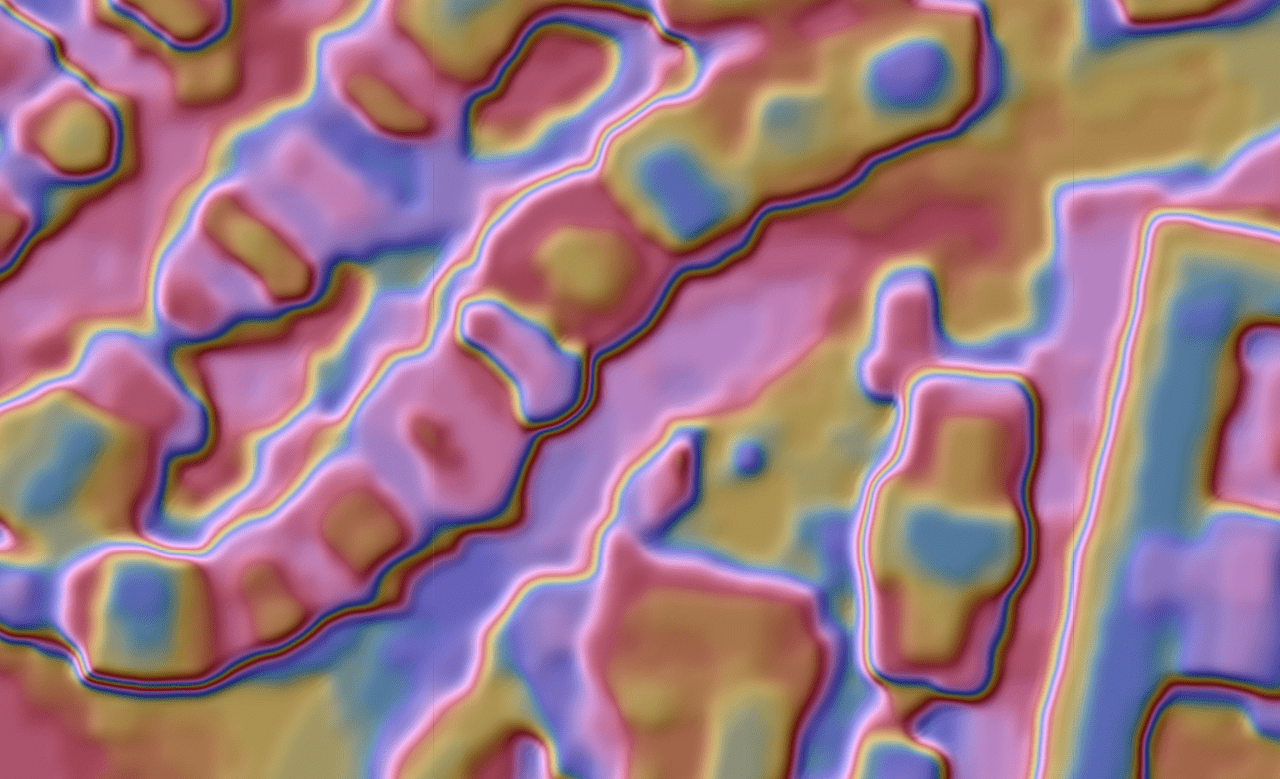}}
        \end{tabular}

        &

        \hspace*{-0.015\textwidth}
        \begin{tikzpicture}
            \node (arrow) {$\rightarrow$};
        \end{tikzpicture}

        &

        \begin{tabular}{c}
            \subcaptionbox{Output nadir RGB\label{fig:subfig3}}{\includegraphics[width=0.29\textwidth]{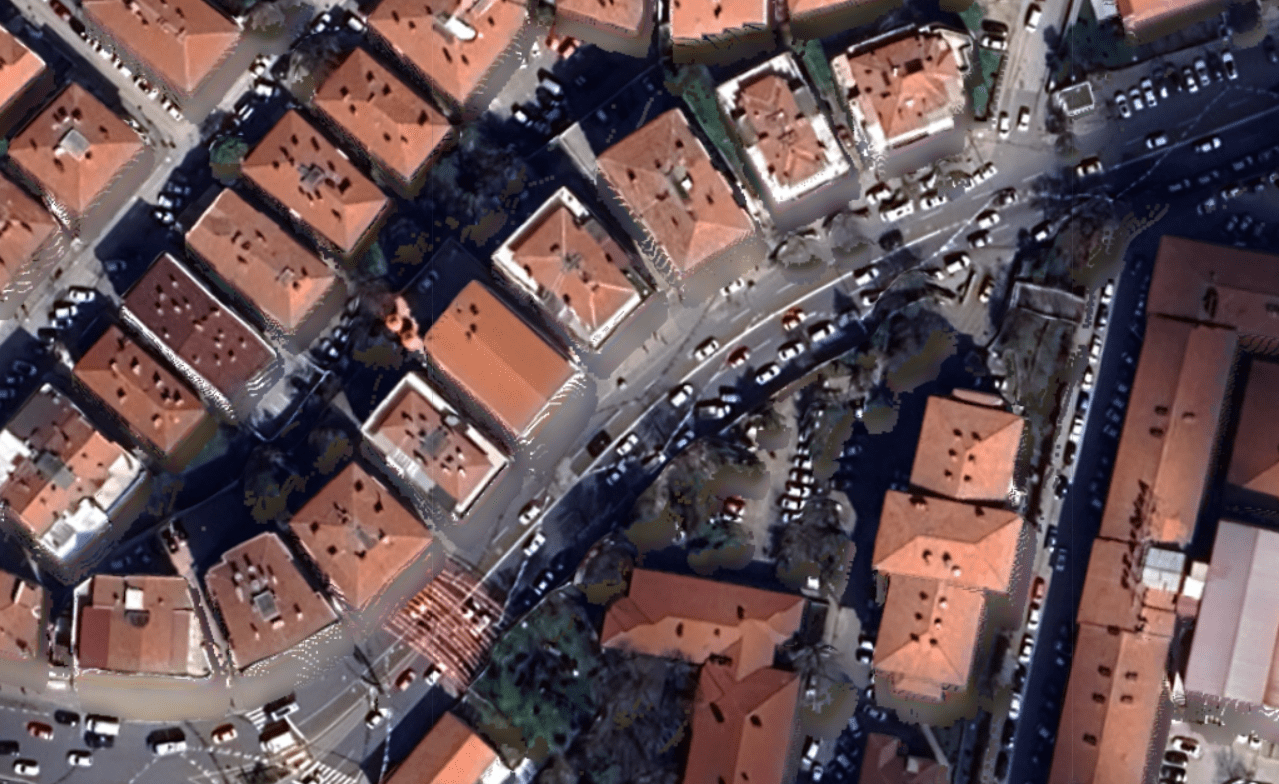}} \\
            + \\
            \subcaptionbox{Output nadir DSM\label{fig:subfig4}}{\includegraphics[width=0.29\textwidth]{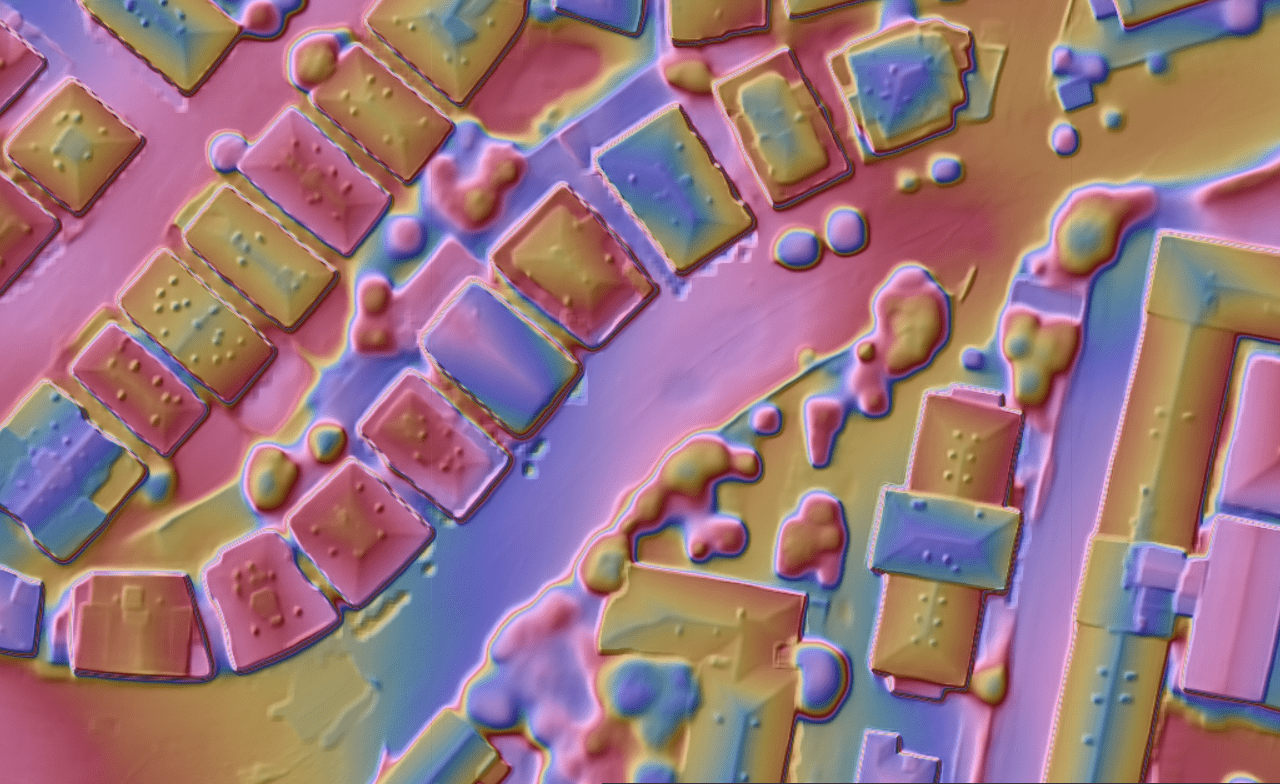}}
        \end{tabular}

        &

        \hspace*{-0.015\textwidth}
        \begin{tikzpicture}
            \node (plus) {+};
        \end{tikzpicture}

        &

        \subcaptionbox{Output nadir roof segments\label{fig:subfig5}}{\includegraphics[width=0.29\textwidth]{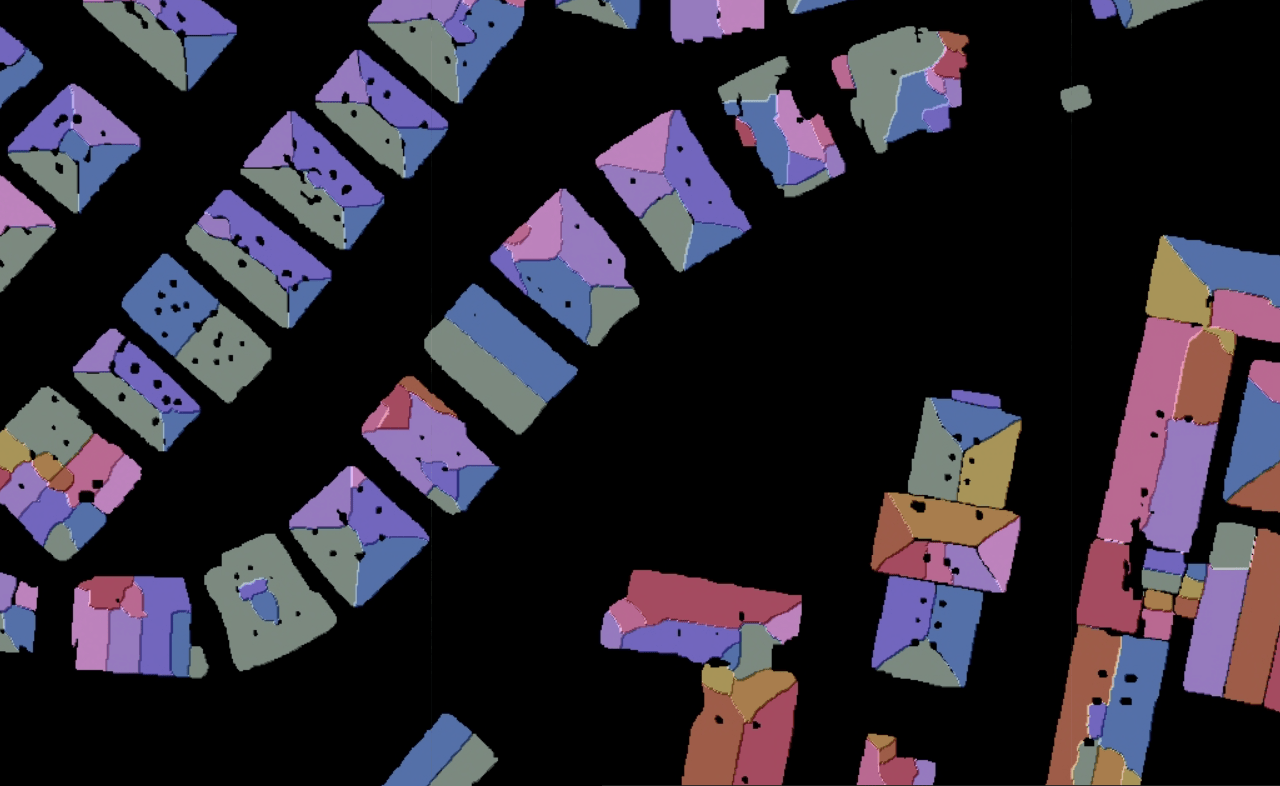}}
    \end{tabular}
    \caption{Overview of inputs and outputs from our ML models. Both DSMs are visualized with a hillshade visualizer. A few sides of buildings are highlighted in the off-nadir satellite RGB with red ovals to emphasize the off-nadir nature of the image. Location: Ankara, Turkey.}
    \label{fig:overview}
\end{figure}
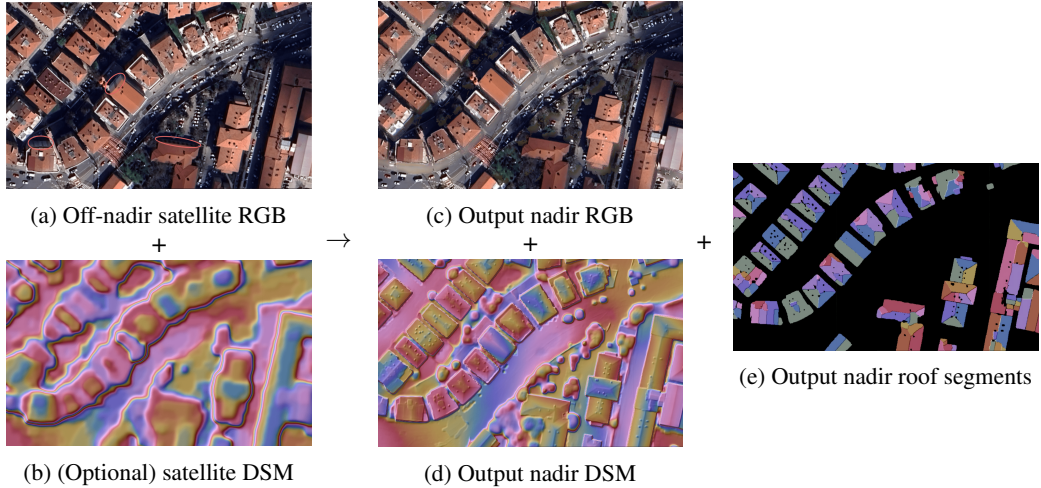

Recent advances in DSM estimation such as \citep{panagiotou2020generating}, \citep{stucker2022resdepth}, \citep{saket2019unet}, have made significant strides, but often overlook fine-grained roof geometry detail which is crucial for solar design. While NeRF-based approaches \citep{roger2022satnerf} offer compelling 3D reconstructions, their multi-view requirement limits scalability. Similarly, existing roof segmentation techniques such as \citep{chen2019aerial}, while effective in certain contexts, often fall short when dealing with satellite inputs. This paper introduces solutions aimed at enhancing DSM estimation and achieving precise instance-level roof segmentation from satellite imagery, thereby facilitating effective solar design.

\section{Data}

\subsection{Inputs, labels and pre-processing}
Inputs comprise off-nadir satellite (Pleiades Neo \citep{pleiadesneo}) RGB at 30cm resolution along with optional photogrammetry derived DSMs and DTMs (plane sweep stereo \citep{collins1996a} with graph cut optimization \citep{boykov2001fast}) computed from a stack of satellite imagery wherever available. These photogrammetry-derived DSMs (even when present) often lack the necessary roof detail required, so we propose enhancements on top.

Labels comprise high quality nadir aerial RGB and corresponding DSMs + DTMs computed via photogrammetry for the DSM estimation task. We compute building instances from the imagery using the a high-quality building detection model \citep{sirko2021continental}. We then run graph cut \citep{boykov2001fast} on the DSMs within each building to produce (somewhat noisy) roof segment labels \citep{goroshin2023estimating} for the roof segment prediction task. We then use simple geometry to reproject the labels into the view frame of each satellite image (see section \ref{sec:appendix_reprojection}). Lastly we compute a few masks: Building presence/consistency masks are computed by comparing satellite and label imagery detected buildings. Similarly, roof segment consistency masks are generated by comparing label roof segments and building instances. These masks are used to filter areas of disagreement. (see section \ref{sec:appendix_masking}).

\subsection{Dataset}
The final dataset is constructed by pairing processed inputs with two sets of labels: off-nadir labels aligned with the input view and nadir labels. This results in \textasciitilde1.1M datapoints in total, with \textasciitilde275k containing satellite DSMs/DTMs as well (referred to as "RGB+DSM" here onwards), and the remaining \textasciitilde860k without (referred to as "RGB only" here onwards). All the inputs + labels are re-sampled to a 25cm resolution in UTM projection. Each datapoint consists of images of 1024x1024 pixels. Spatial distributions of the dataset splits are visualized in section \ref{sec:appendix_geo_distribution}. We generate 60:20:20 train:val:test splits at the city level (i.e. the splits consist of different cities) and then sub-sample the validation and test splits to ensure an equal per-country sampling of imagery.

In addition we collect a human annotated dataset of 1647 tiles where roof segments were annotated in the off-nadir RGB satellite imagery (see section \ref{sec:appendix_human_labels}). We use these labels for additional validation to complement our noisier graph cut segmentation.

\section{Methods}

\begin{figure}[h!]
    \centering
    \begin{tabular}{c|c c}
        \subcaptionbox{Off-nadir satellite RGB\label{fig:subfig1}}{\includegraphics[width=0.3\textwidth]{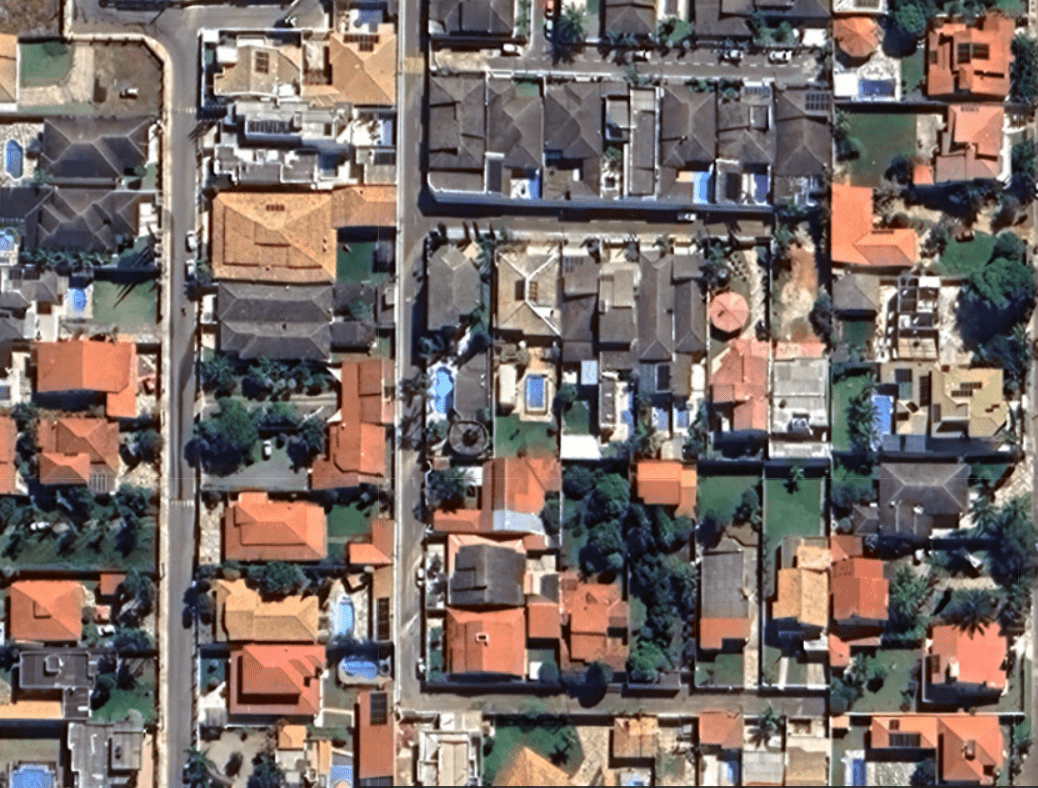}} &
        \subcaptionbox{Output nadir DSM\label{fig:subfig2}}{\includegraphics[width=0.3\textwidth]{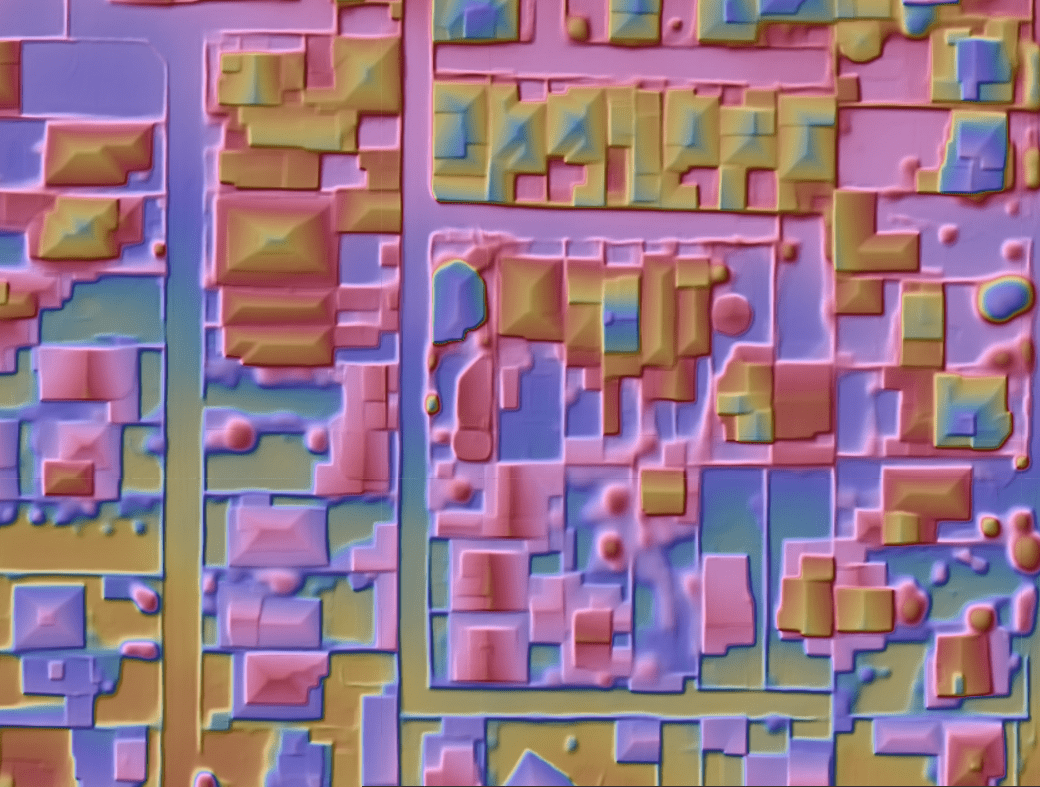}} &
        \subcaptionbox{Output roof segments\label{fig:subfig3}}{\includegraphics[width=0.3\textwidth]{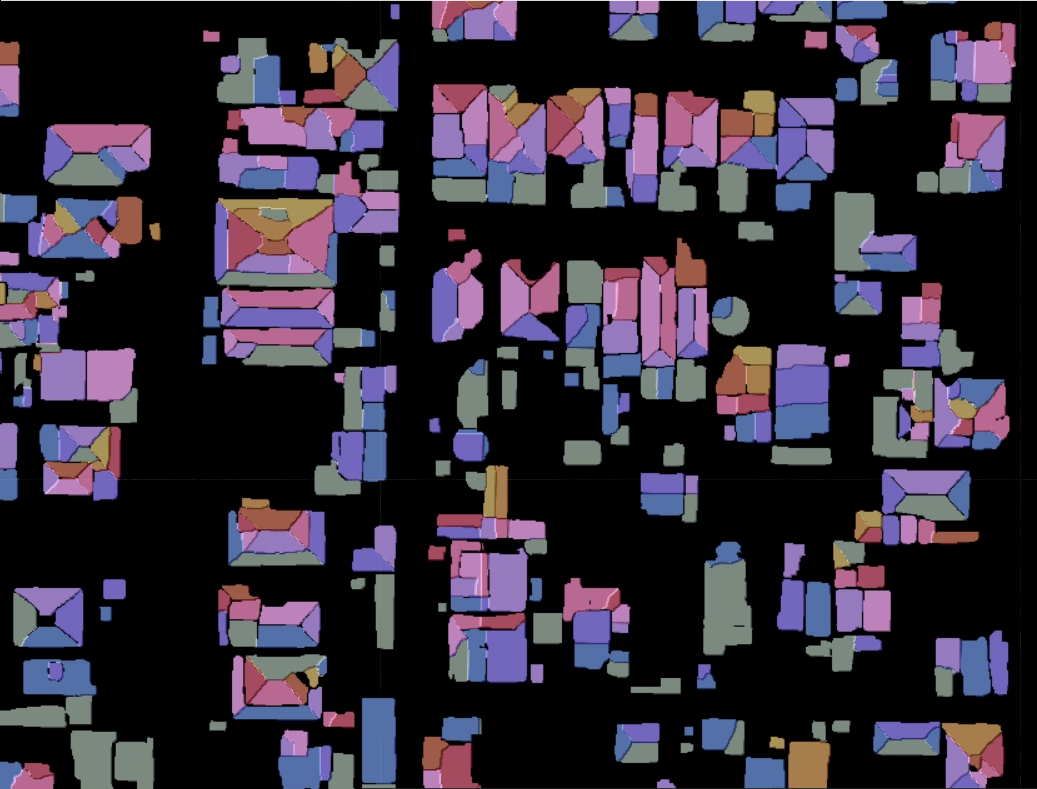}} \\

        \subcaptionbox{(Optional) satellite DSM\label{fig:subfig4}}{\includegraphics[width=0.3\textwidth]{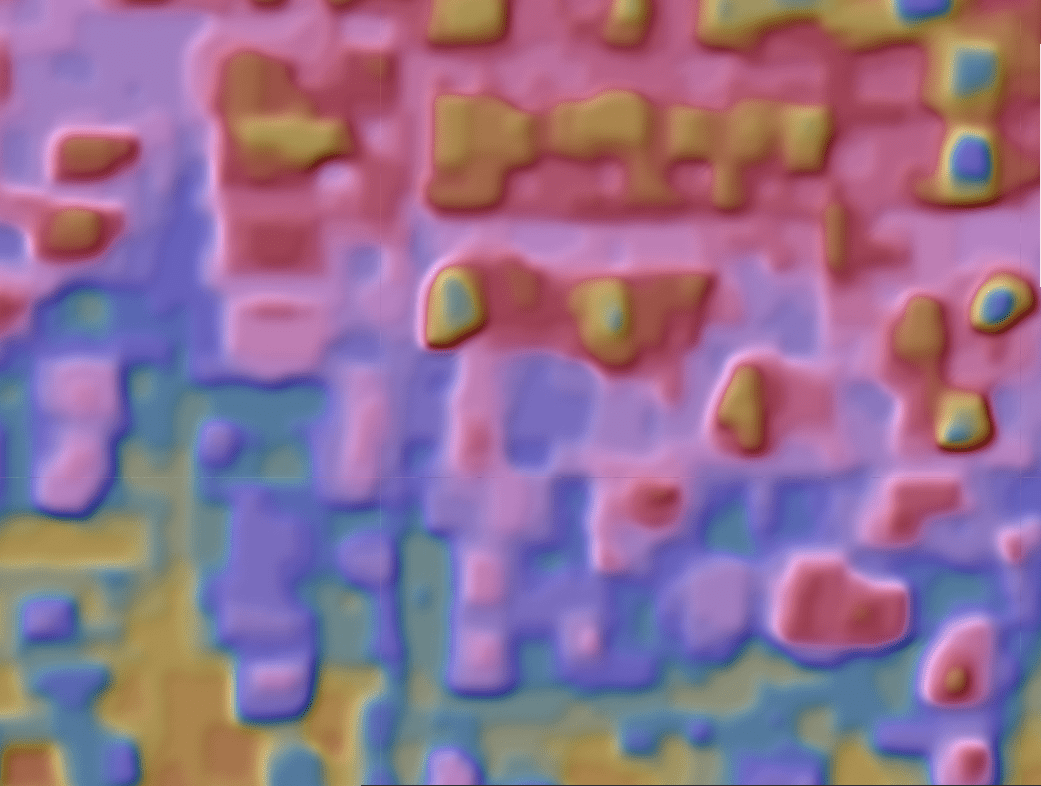}} &
        \subcaptionbox{Output flux maps\label{fig:subfig5}}{\includegraphics[width=0.3\textwidth]{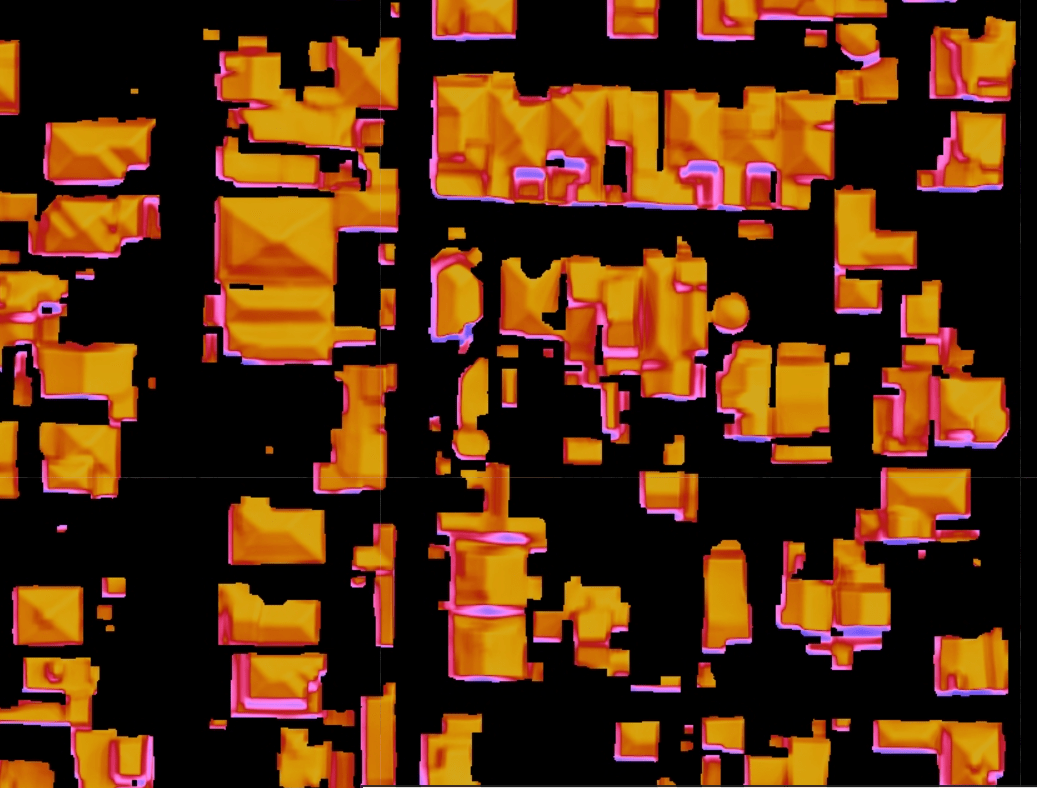}} &
        \subcaptionbox{Output panel placement\label{fig:subfig6}}{\includegraphics[width=0.3\textwidth]{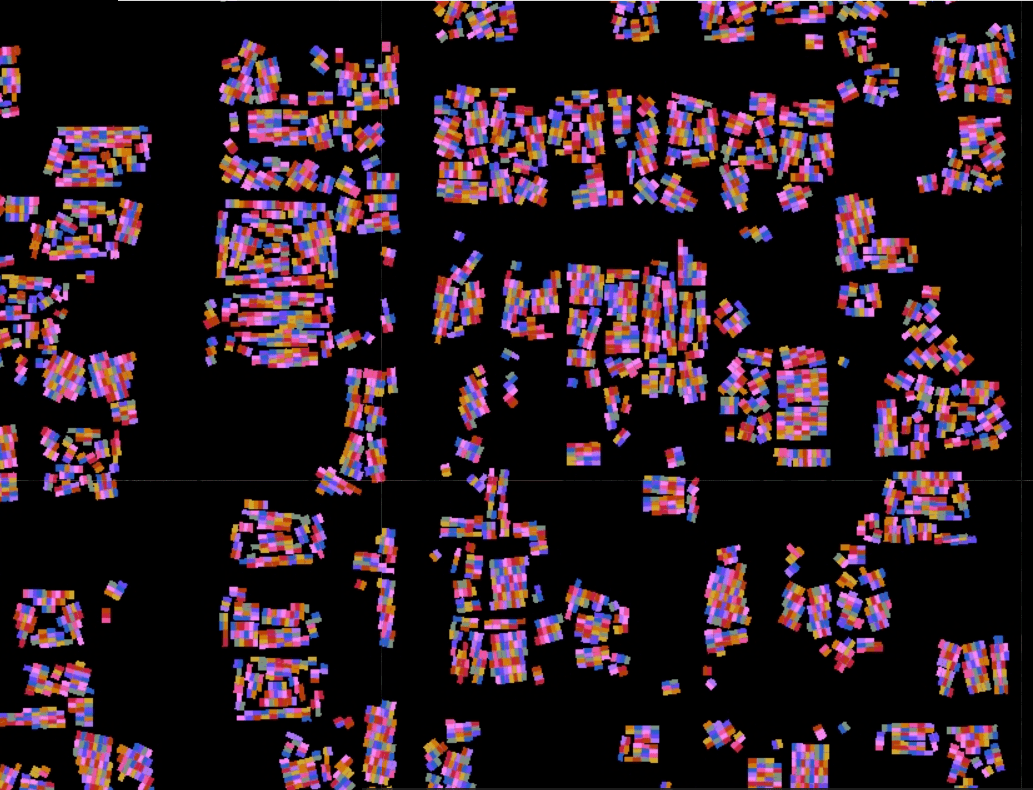}}
    \end{tabular}
    \caption{Sample inputs and outputs from the Satellite Solar API pipeline. All outputs are nadir. Location: Brasilia, Brazil.}
    \label{fig:solar_api_outputs}
\end{figure}

We train base and refinement models which take in off-nadir satellite imagery as inputs and produce nadir outputs which are consumed by the Satellite Solar API pipeline.

\subsection{Base model}

The base model processes off-nadir satellite RGB, optionally incorporating a photogrammetry derived height map (DSM minus DTM) and satellite viewing angles (elevation and azimuth) to generate enhanced off-nadir height maps and roof segment instances. Subsequently, we reproject the off-nadir satellite imagery (see section \ref{sec:appendix_reprojection}), enhanced height maps and roof segments to a nadir view using the enhanced height maps.

Similar to \citep{goroshin2023estimating}, we use a U-Net \citep{ronneberger2015u} styled architecture employing a Swin Transformer \citep{liu2021swin} encoder for feature extraction, followed by a convolutional up-sampling decoder (see section \ref{sec:appendix_model_details} for details and hyper-parameters). Two prediction heads are added on top of the decoder: one dedicated to height map regression, the other to affinity mask prediction for roof segment delineation as described in \citet{goroshin2023estimating}.

The model is trained using an L1 loss to minimize height map discrepancies, a Sobel gradient \citep{kanopoulos1988design} loss to achieve smooth gradients and capture finer roof details (crucial for panel placement later on), and an affinity mask loss for roof instance segmentation \citep{goroshin2023estimating}. Performance is evaluated using L1 height map error, roof segment intersection over union, and pitch/azimuth errors computed from per-pixel surface normals (obtained from the height maps) averaged within each segment.

To enhance the accuracy and alignment of labels used in losses and metrics with respect to the satellite inputs, we employ various masking techniques (discussed in detail in section \ref{sec:appendix_masking}).

\subsection{Refinement model}

This stage takes as input nadir RGB imagery and height maps (containing occlusions due to reprojection) from the base model, and produces refined nadir RGB, height maps, and roof segments devoid of such artifacts.

The model architecture largely mirrors that of the base model, with the key addition of an RGB prediction head alongside the height map and roof segment heads. An L2 loss is employed between aerial RGB and the model's predicted RGB. Model predicted RGB is then utilized to fill any occlusions present in the original nadir satellite RGB from the base model (more details in section \ref{sec:appendix_refinement_rgb}).

To derive the final refined DSMs, we combine model-predicted height maps with available low-quality satellite DTMs. In the absence of such DTMs, we incorporate re-sampled NASA DEMs \citep{crippen2016nasadem} originally at 30m resolution.

\subsection{Satellite Solar API pipeline}

The Solar API pipeline, as detailed in \citet{goroshin2023estimating}, estimates solar potential and panel layouts (visualized in Figure \ref{fig:solar_api_outputs}) by utilizing nadir RGB imagery, DSMs, and roof segments as input. It runs model inference on overlapping tiles, and stitches the output seamlessly using weighted kernels. It then uses ray-tracing to generate building-level solar flux estimates and proposes optimal solar panel placement configurations.

We assess end-to-end performance through metrics such as Mean Absolute Percentage Error (MAPE) and MAPE@5kW \citep{goroshin2023estimating} on the Solar API pipeline outputs.

\section{Results and future work}

Quantitative evaluation of our base + refinement models combined, using metrics defined in the preceding section, is presented in Table \ref{metrics_summary_table}. We divide our metrics into results from the "RGB only" and "RGB+DSM" datasets and further split the results for the latter into RGBH (RGB + height map) inputs and RGB (no heightmap) inputs to probe the performance impact of losing the input height map on a consistent dataset. These results suggest that while height map inputs improve overall DSM accuracy, they are not essential for achieving high-quality roof geometry and segmentation results.

\begin{table}
  \caption{Validation and test results for the combined base + refinement model. Columns left to right: Split, dataset, input channels used by the model, overall height map MAE (mean absolute error), height map MAE over buildings only, graph-cut (GC) based roof pitch/azimuth errors, graph-cut based IoU (intersection over union), human-labels based IoU.}
  \label{metrics_summary_table}
  \centering
  \begin{tabular}{lcccccccc}
    \toprule
     Split & Dataset & Input & \makecell{Overall \\ MAE} & \makecell{Building \\ MAE} & \makecell{GC \\ pitch \\ error} & \makecell{GC \\ azimuth \\ error} & \makecell{GC \\ segment \\ IOU} & \makecell{Human \\ labels \\ segment \\ IOU} \\
    \midrule
    \multirow{3}{*}{Val} & RGB+DSM & RGBH & 1.51m & 1.17m & 4.73$^\circ$ & 13.2$^\circ$ & 54.2\% & 56.0\% \\
    & RGB+DSM & RGB & 1.65m & 1.33m & 4.81$^\circ$ & 13.7$^\circ$ & 53.6\% & 56.2\% \\
    & RGB only & RGB & 1.41m & 1.07m & 5.08$^\circ$ & 12.3$^\circ$ & 52.3\% & - \\
    \midrule
    \multirow{3}{*}{Test} & RGB+DSM & RGBH & 1.24m & 1.03m & 4.83$^\circ$ & 13.1$^\circ$ & 52.4\% & - \\
    & RGB+DSM & RGB & 1.33m & 1.16m & 4.97$^\circ$ & 14.0$^\circ$ & 51.8\% & - \\
    & RGB only & RGB &1.26m & 0.92m & 5.13$^\circ$ & 10.2$^\circ$ & 54.7\% & - \\
    \bottomrule
  \end{tabular}
\end{table}

In addition to these top-level metrics, in the appendix we include insights into per-country performance (section \ref{sec:appendix_country_level_analysis}, where aerial data was available), end-to-end metrics from the Solar API pipeline (section \ref{sec:appendix_e2e_metrics}) and ablations and sensitivity analyses studying the impact of key factors (section \ref{sec:appendix_ablations}, including dataset size and masking).

Future work will focus on refining solar potential estimates by addressing challenges such as roof obstacle detection, roof material/type detection, existing solar panel identification and roof segment fine-tuning with human labels.

\begin{ack}

We would like to thank John C. Platt, Artem Zholus, Christopher Schmidt, Jordan Raisher, Saleem V. Groenou, Dana Kurnaiwan, Juliet Rothenberg and Courtney Maimon for their valuable insights and suggestions.

\end{ack}

\medskip

\small

\bibliography{tackling_climate_workshop}
\bibliographystyle{plainnat}

%%%%%%%%%%%%%%%%%%%%%%%%%%%%%%%%%%%%%%%%%%%%%%%%%%%%%%%%%%%%

% Stop any content from the appendix being flowed into the
% main body of the paper and sabotaging our page limit.
\clearpage

\appendix

\section{Data}

\subsection{Human labeled roof segments}
\label{sec:appendix_human_labels}

\begin{figure}[h!]
    \centering
    \begin{tabular}{cc}
        \subcaptionbox{Off-nadir satellite RGB\label{fig:subfig1}}{\includegraphics[width=0.4\textwidth]{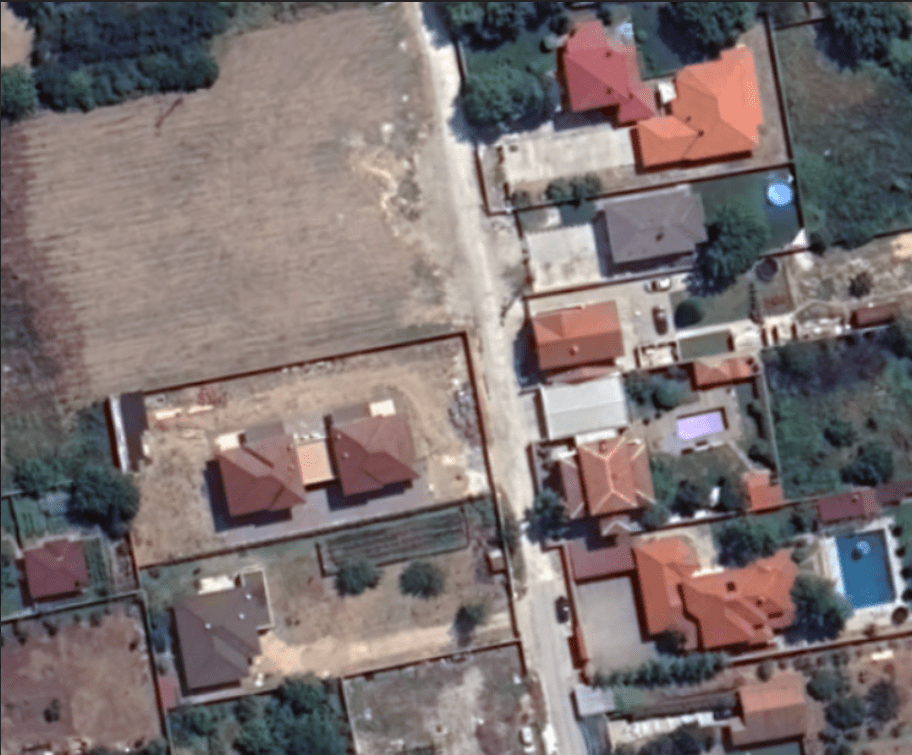}} &
        \subcaptionbox{Human labeled roof segments\label{fig:subfig6}}{\includegraphics[width=0.4\textwidth]{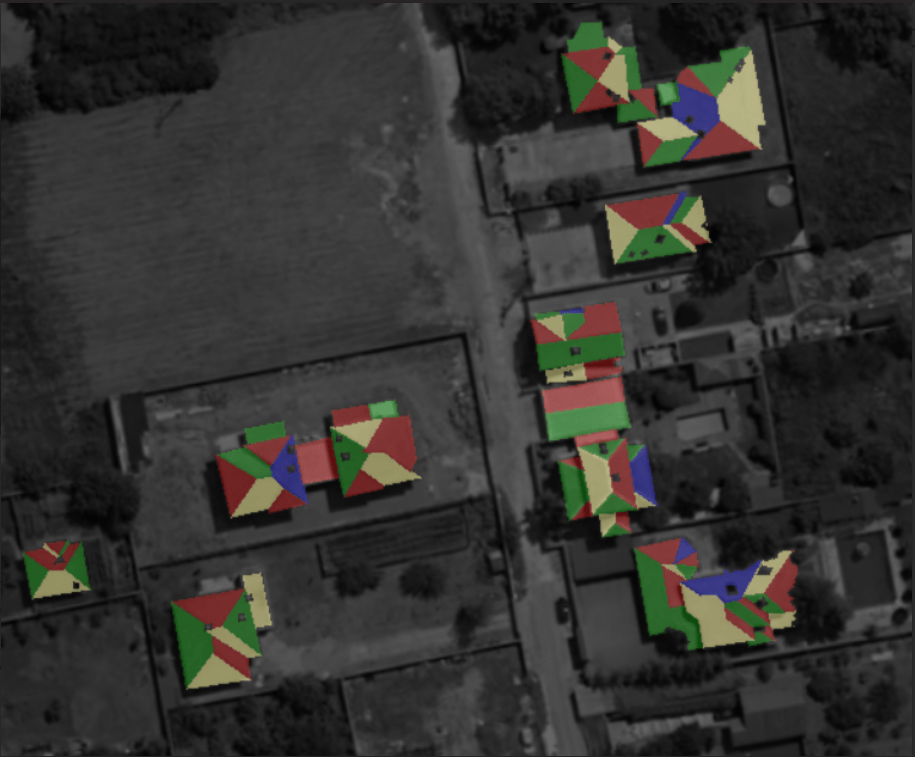}} \\
    \end{tabular}
    \caption{Human labeled roof segments visualization.}
    \label{fig:human_labels}
\end{figure}

Recognizing the inherent noise in our graph cut roof segment labels, we add human-annotated labels to our evaluation process to produce more reliable roof segmentation metrics. Each annotator is presented with a satellite RGB scene and is tasked with labeling roof segments. Due to the high building density in each scene, we ask annotators to label as many buildings as possible in a contiguous area in the center of each image \citep{wang2022unlocking}. To aid in discerning finer details, annotators can reference corresponding aerial RGB and DSM data alongside the satellite RGB. An example of human annotated labels are shown in Figure \ref{fig:human_labels}.

\subsection{Geometry based reprojection}
\label{sec:appendix_reprojection}

\begin{figure}[h!]
    \centering
    \begin{tabular}{c c|c c}
        \subcaptionbox{Aerial nadir RGB\label{fig:subfig1}}{\includegraphics[width=0.3\textwidth]{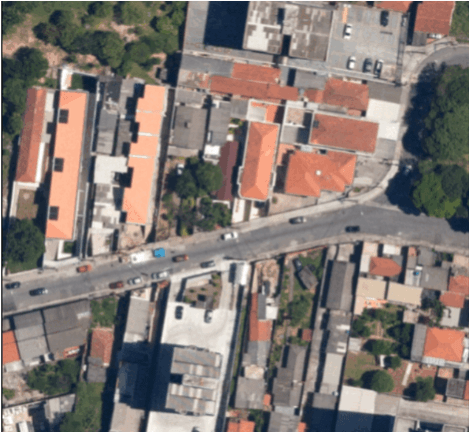}} &
        \subcaptionbox{Reprojected aerial RGB\label{fig:subfig2}}{\includegraphics[width=0.3\textwidth]{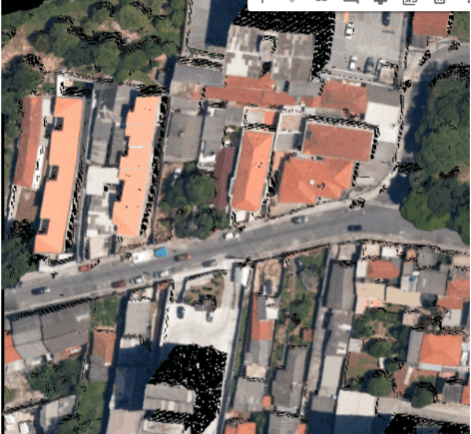}} &
        \multirow{2}{*}{\makebox[0.3\textwidth]{\subcaptionbox{Reference satellite RGB\label{fig:subfig3}}{\includegraphics[width=0.3\textwidth]{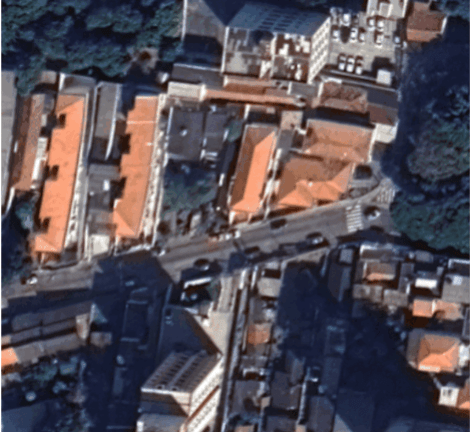}}}} \\\\

        \subcaptionbox{Aerial nadir DSM\label{fig:subfig4}}{\includegraphics[width=0.3\textwidth]{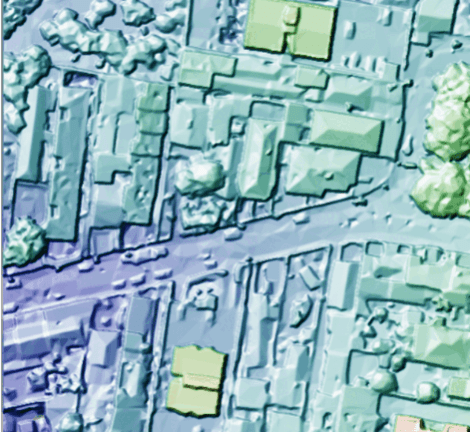}} &
        \subcaptionbox{Reprojected aerial DSM\label{fig:subfig5}}{\includegraphics[width=0.3\textwidth]{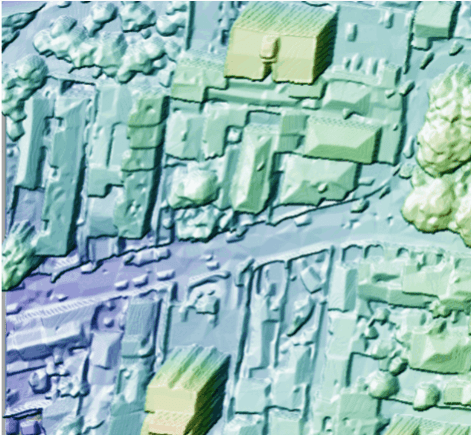}} &
    \end{tabular}
    \caption{Geometry-based reprojection.}
    \label{fig:reprojection}
\end{figure}

We opt for a simplified geometric reprojection approach that is efficient and suits our use case. Predicated on the assumptions of an infinitely distant satellite and parallel ground rays, we derive the following set of reprojection equations. Sample reprojected outputs are shown in Figure \ref{fig:reprojection}.

Given input satellite angles (elevation, azimuth), we obtain y and x angles for reprojection as,

\[angle_y = \arctan{(cos(azimuth) / tan(elevation))}\]
\[angle_x = \arctan{(sin(azimuth) / tan(elevation))}\]

We reproject each pixel individually using the angles computed above, in ascending order of original height. If multiple pixels map to the same target location, we retain the value of the highest original pixel to handle occlusion.

\[reprojected\_y_i = round(y_i + (height_i / spatial\_resolution) * \tan{angle_y})\]
\[reprojected\_x_i = round(x_i + (height_i / spatial\_resolution) * \tan{angle_x})\]

When reprojecting height maps from nadir to off-nadir in particular, we make a small modification to reproject entire sides of buildings and not just leave them as masked out pixels. For each input pixel, we get the neighbouring pixel with the smallest height (\(h\_base_i\)) and then reproject the current pixel multiple times starting from \(h\_base_i\) till \(h_i\) at 1m intervals. This produces smooth gradients on sides of buildings which are useful approximations for training models.

\subsection{Distribution of train/val/test dataset splits}
\label{sec:appendix_geo_distribution}

\begin{figure}
  \centering
  \includegraphics[width=\textwidth]{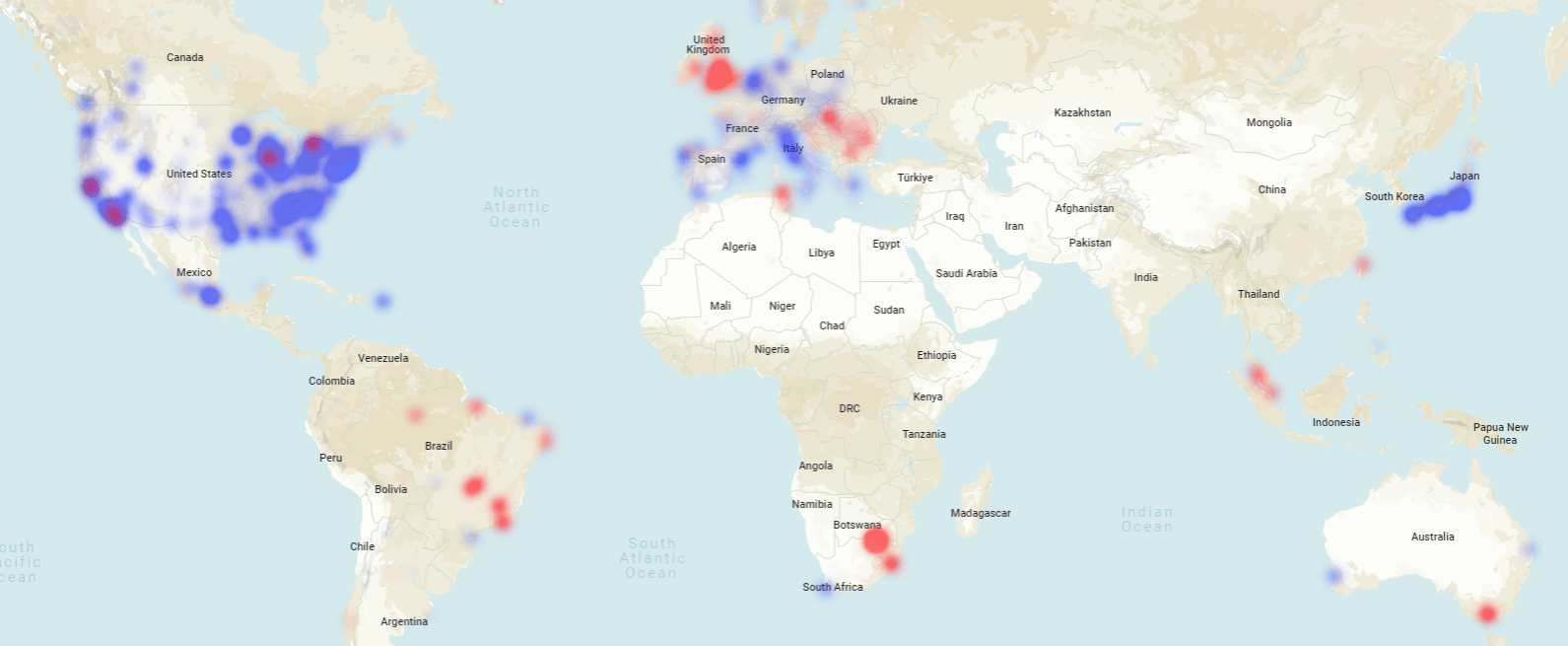}
  \includegraphics[width=\textwidth]{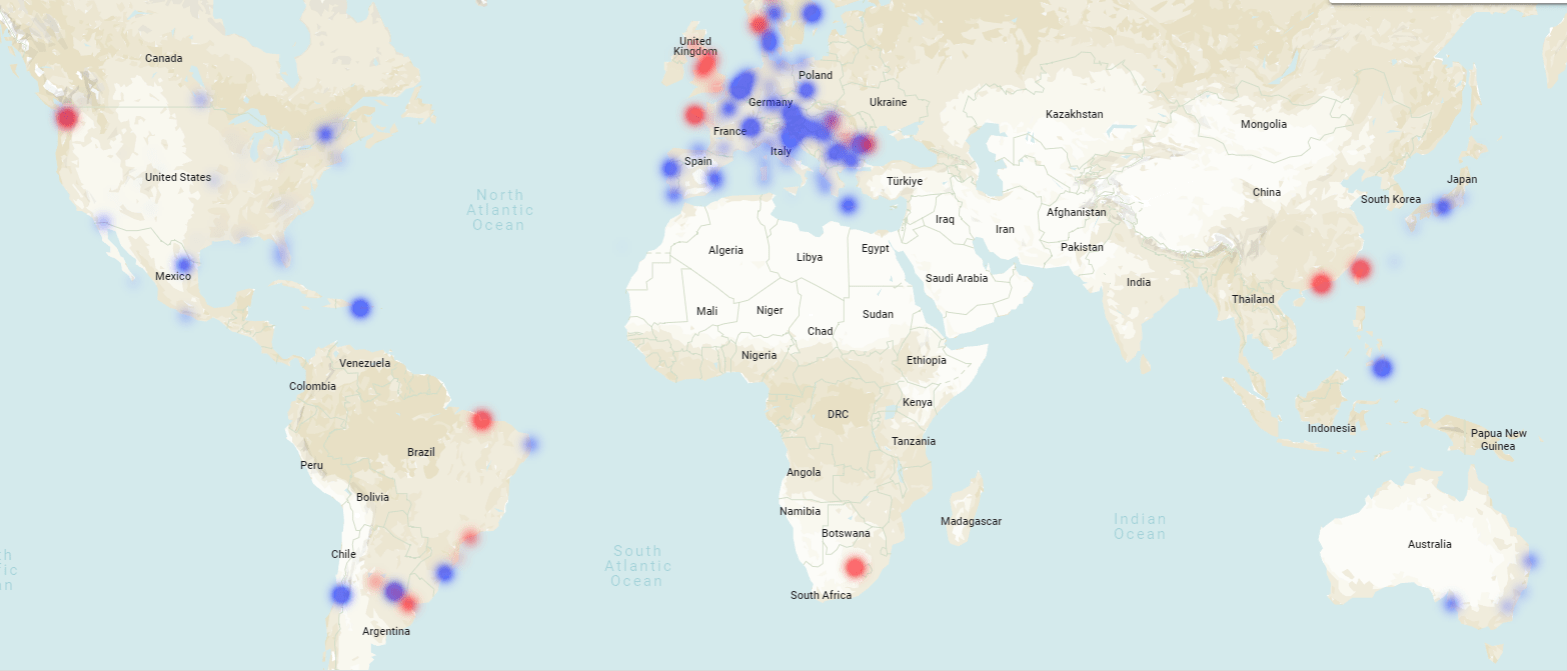}
  \includegraphics[width=\textwidth]{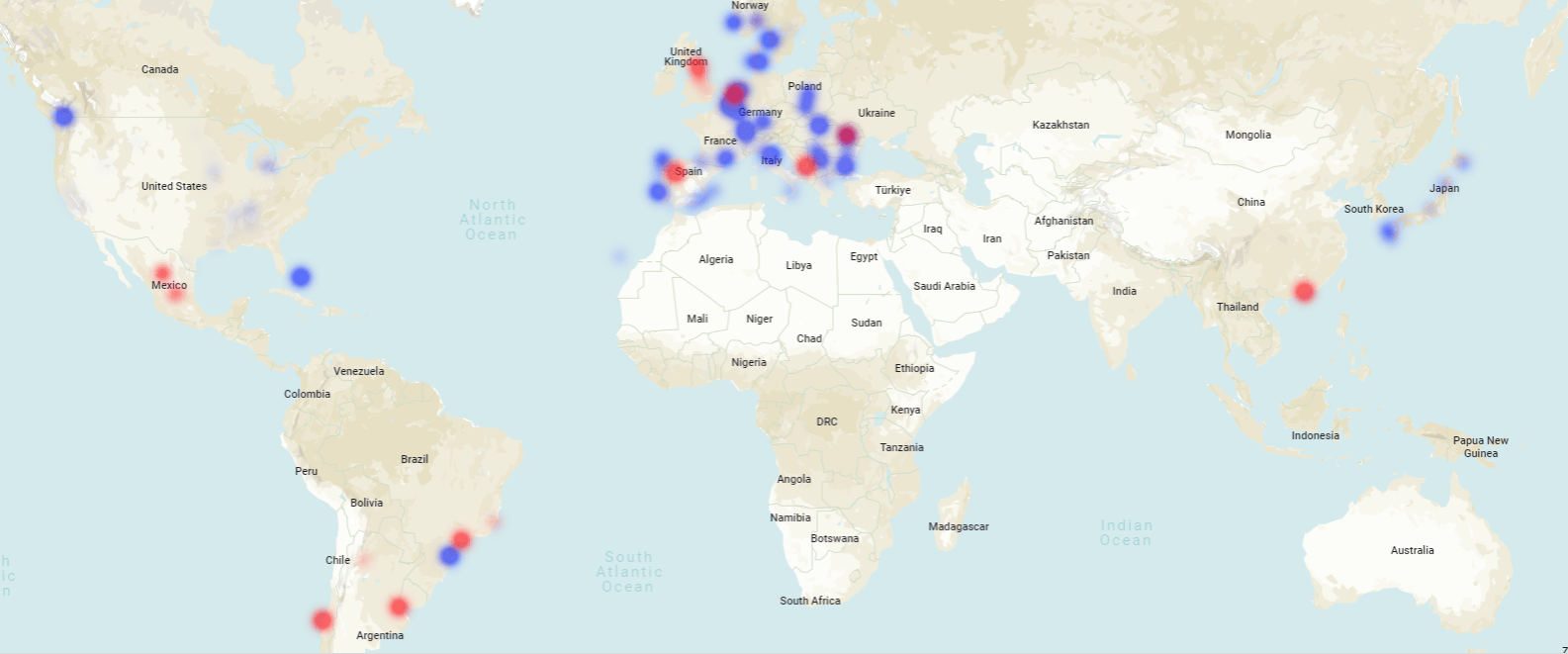}
  \caption{Geographical distribution of train, validation and test splits (in order). Red represents RGB+DSM data and blue represents RGB only data.}
  \label{fig:data_distribution}
\end{figure}

Figure \ref{fig:data_distribution} illustrates the geographic distributions of our train, validation, and test splits. To ensure greater geographic diversity in the validation and test sets, we sub-sample the data to maintain roughly equal representation from each country. The training set utilizes all available data.

Note that RGB+DSM and RGB only data are mutually exclusive.

\section{Modeling}

\subsection{Masking techniques for loss/metrics}
\label{sec:appendix_masking}

The quality of training data significantly impacts the performance of our models. To mitigate the influence of noisy or erroneous data during training (loss computation) and evaluation (metrics), we implement the following masking schemes:

\begin{enumerate}
    \item Temporal building mismatch masking: Temporal discrepancies between aerial DSM labels and satellite inputs can lead to inconsistencies due to building demolitions or new constructions. We mask out pixels where the high-quality building masks \citep{sirko2021continental} derived from satellite and aerial RGB imagery disagree with each other, thereby excluding potentially changed areas.
    \item Roof segment masking: Graph cut roof segment labels from aerial imagery can be noisy and may not cover buildings fully. Utilizing high-quality building instances from satellite imagery, we mask out buildings where less than 50\% of their area is covered by graph cut roof segments. This helps filter out potential graph cut errors, as roof segments are generally expected to encompass a majority of the building area.
\end{enumerate}

\subsection{Model details + Hyper-parameters}
\label{sec:appendix_model_details}

Our model adopts a U-Net \citep{ronneberger2015u} style architecture, featuring a SWIN-B \citep{liu2021swin} encoder for feature extraction and a convolutional up-sampling decoder. The decoder comprises three up-sampling stages, each consisting of two convolutional blocks (each block consisting of convolution, normalization and activation layers) with skip connections. We attach two prediction heads to the decoder output, each comprising of a single convolution layer. We employ SELU activations \citep{klambauer2017self} throughout the model.

We use the following hyper-parameters:
\begin{itemize}
\item Learning rate: 0.0003
\item Weight decay: 1e-07
\item Training batch size: 1024
\item Input/label size: 1024x1024 (randomly cropped to 512x512 during training)
\item Evaluation tile size: 1024x1024
\item Total training steps: 75000
\item Learning rate schedule: Warm-up + cosine decay (warm-up steps = 7,500)
\end{itemize}

The following loss weighting schemes are also applied:
\begin{itemize}
\item Affinity mask losses: Up-weighted by 5x to align with the scale of DSM losses.
\item DSM and gradient losses within buildings: Up-weighted by 5x to prioritize rooftop geometry.
\item Vegetation: Implicitly down-weighted to mitigate the impact of noisy vegetation labels due to seasonal variations and temporal gaps between labels and input data.
\end{itemize}

\subsection{Refinement RGB post-processing}
\label{sec:appendix_refinement_rgb}

\begin{figure}[h!]
    \centering
    \begin{tikzpicture}
        \node (img1) at (-2.5, 0) {
            \begin{minipage}{0.23\textwidth}
                \centering
                \includegraphics[width=\textwidth, height=\textwidth]{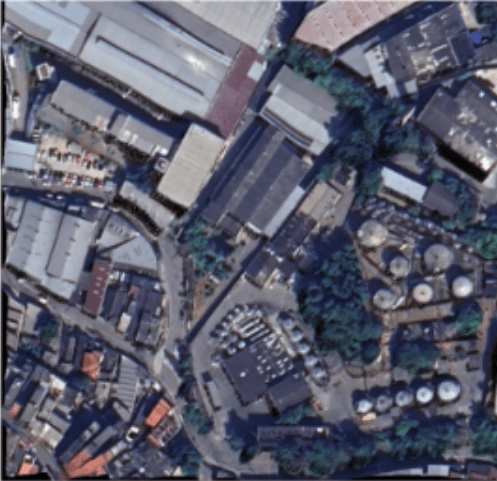}
                \subcaption{RGB with occlusions}
            \end{minipage}
        };
        
        \node at (-0.75, 0.1) {$+$};
        
        \node (img2) at (1, 0) {
            \begin{minipage}{0.23\textwidth}
                \centering
                \includegraphics[width=\textwidth, height=\textwidth]{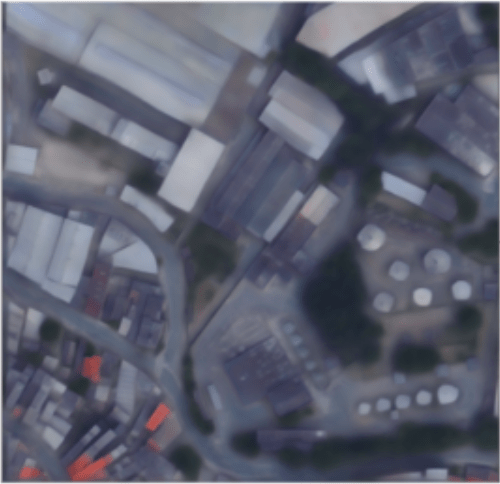}
                \subcaption{Model predicted RGB}
            \end{minipage}
        };
        
        \node at (2.75, 0.1) {$+$};
        
        \node (img3) at (4.5, 0) {
            \begin{minipage}{0.23\textwidth}
                \centering
                \includegraphics[width=\textwidth, height=\textwidth]{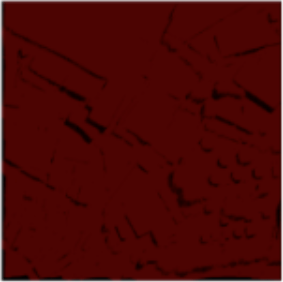}
                \subcaption{Occlusions}
            \end{minipage}
        };
        
        \node at (6.25, 0.1) {$\rightarrow$};
        
        \node (result) at (8.1, 0) {
            \begin{minipage}{0.23\textwidth}
                \centering
                \includegraphics[width=\textwidth, height=\textwidth]{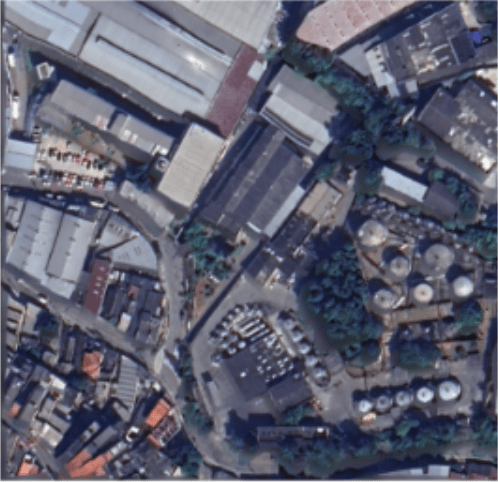}
                \subcaption{Infilled RGB}
            \end{minipage}
        };
        
    \end{tikzpicture}
    \caption{Generating the final infilled nadir RGB output.}
    \label{fig:refinement_rgb_output}
\end{figure}
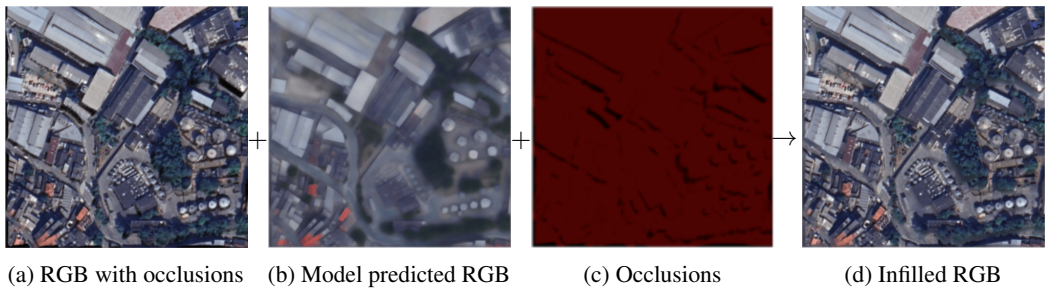

Occlusions in geometry reprojected satellite imagery appear as distracting black regions. We address this by in-filling these areas with a blurry model-predicted RGB as outlined in Figure \ref{fig:refinement_rgb_output}. This maintains visual clarity without obscuring the distinction between real and synthetic data.

\section{Results}

\subsection{Ablations}
\label{sec:appendix_ablations}

\begin{table}
  \caption{Masking ablation results on the RGB+DSM validation split with RGBH inputs.}
  \label{table:masking_ablation_results}
  \centering
  \begin{tabular}{lcccccc}
    \toprule
    Method & \makecell{Overall \\ MAE} & \makecell{Building \\ MAE} & \makecell{GC \\ pitch \\ error} & \makecell{GC \\ azimuth \\ error} & \makecell{GC \\ segment \\ IOU} & \makecell{Human \\ labels \\ segment \\ IOU} \\
    \midrule
    Baseline & \textbf{1.42m} & \textbf{1.38m} & \textbf{4.36$^\circ$} & 18.6$^\circ$ & \textbf{53.7\%} & \textbf{55.4\%} \\
    \makecell[l]{No temporal building \\ mismatch masking} & 1.46m & 1.44m & 4.5$^\circ$ & 18.4$^\circ$ & 53.2\% & 53.9\% \\
    \makecell[l]{No roof \\ segment masking} & 1.47m & 1.51m & 4.38$^\circ$ & 18.9$^\circ$ & 53.3\% & 53.5\% \\
    \makecell[l]{No masking} & 1.47m & 1.47m & 4.55$^\circ$ & \textbf{18.2$^\circ$} & 52.5\% & 51.1\% \\
    \bottomrule
  \end{tabular}
\end{table}

\begin{table}
  \caption{Dataset size ablation results on the RGB+DSM validation split with RGBH inputs.}
  \label{table:dataset_size_ablation_results}
  \centering
  \begin{tabular}{lcccccc}
    \toprule
    Dataset size & \makecell{Overall \\ MAE} & \makecell{Building \\ MAE} & \makecell{GC \\ pitch \\ error} & \makecell{GC \\ azimuth \\ error} & \makecell{GC \\ segment \\ IOU} \\
    \midrule
    1M (baseline) & \textbf{1.42m} & \textbf{1.38m} & \textbf{4.36$^\circ$} & \textbf{18.6$^\circ$} & \textbf{53.7\%} \\
    360k & 1.45m & 1.39m & 4.49$^\circ$ & 19.4$^\circ$ & 53.7\% \\ 
    200k & 1.45m & 1.49m & 4.52$^\circ$ & 19.4$^\circ$ & 53.6\% \\ 
    60k & 1.65m & 1.9m & 4.62$^\circ$ & 20.6$^\circ$ & 53.0\% \\ 
    20k & 1.66m & 1.9m & 4.91$^\circ$ & 21.4$^\circ$ & 51.8\% \\ 
    6k & 1.88m & 2.55m & 5.12$^\circ$ & 23.9$^\circ$ & 50.5\% \\
    \bottomrule
  \end{tabular}
\end{table}

To gain a deeper understanding of the contributions of various modeling components, we conduct two sets of ablation studies.

\begin{enumerate}
    \item Masking ablation: We drop each of our masking techniques one at a time to assess its impact on the overall performance of our models in Table \ref{table:masking_ablation_results}.
    \item Input dataset size ablation: We progressively decrease the size of the input dataset (logarithmically) to examine its effect on model performance in Table \ref{table:dataset_size_ablation_results}.
\end{enumerate}

Note that both these sets of ablation results are on the base model only (opted for faster experimentation since refinement models take longer to train). So these are not comparable with the main results in Table \ref{metrics_summary_table} directly.

We observe (with the marginal exception of azimuth error) that the best model performance is obtained when combining both masking strategies during training, and that the model performance continues to improve with increasing dataset size: with training with 1M datapoints resulting in the best performance.

\subsection{End-to-end MAPE metrics}
\label{sec:appendix_e2e_metrics}

\begin{table}
  \caption{End-to-end performance metrics - MAPE@5kW and MAPE computed across randomly sub-sampled validation and test splits where we include one region from each country.}
  \label{table:e2e_perf_metrics}
  \centering
  \begin{tabular}{lccc}
    \toprule
    Split & MAPE@5kw & MAPE \\
    \midrule
    Validation & 2.6\% & 19.2\% \\
    Test & 2.5\% & 18.3\% \\
    \bottomrule
  \end{tabular}
\end{table}

End-to-end performance metrics are presented in Table \ref{table:e2e_perf_metrics}. Test outputs are obtained by feeding the model-predicted nadir RGBs, DSMs, and roof segments into the Solar API pipeline. Ground truth outputs are derived by directly running the Solar API pipeline on high-quality aerial imagery of the same regions wherever overlap exists. Predicted fluxes and panel placements are then compared against each other to compute performance metrics (as outlined in \citep{goroshin2023estimating}).

\subsection{Country level analysis}
\label{sec:appendix_country_level_analysis}

\begin{figure}
  \centering
  \includegraphics[width=\textwidth]{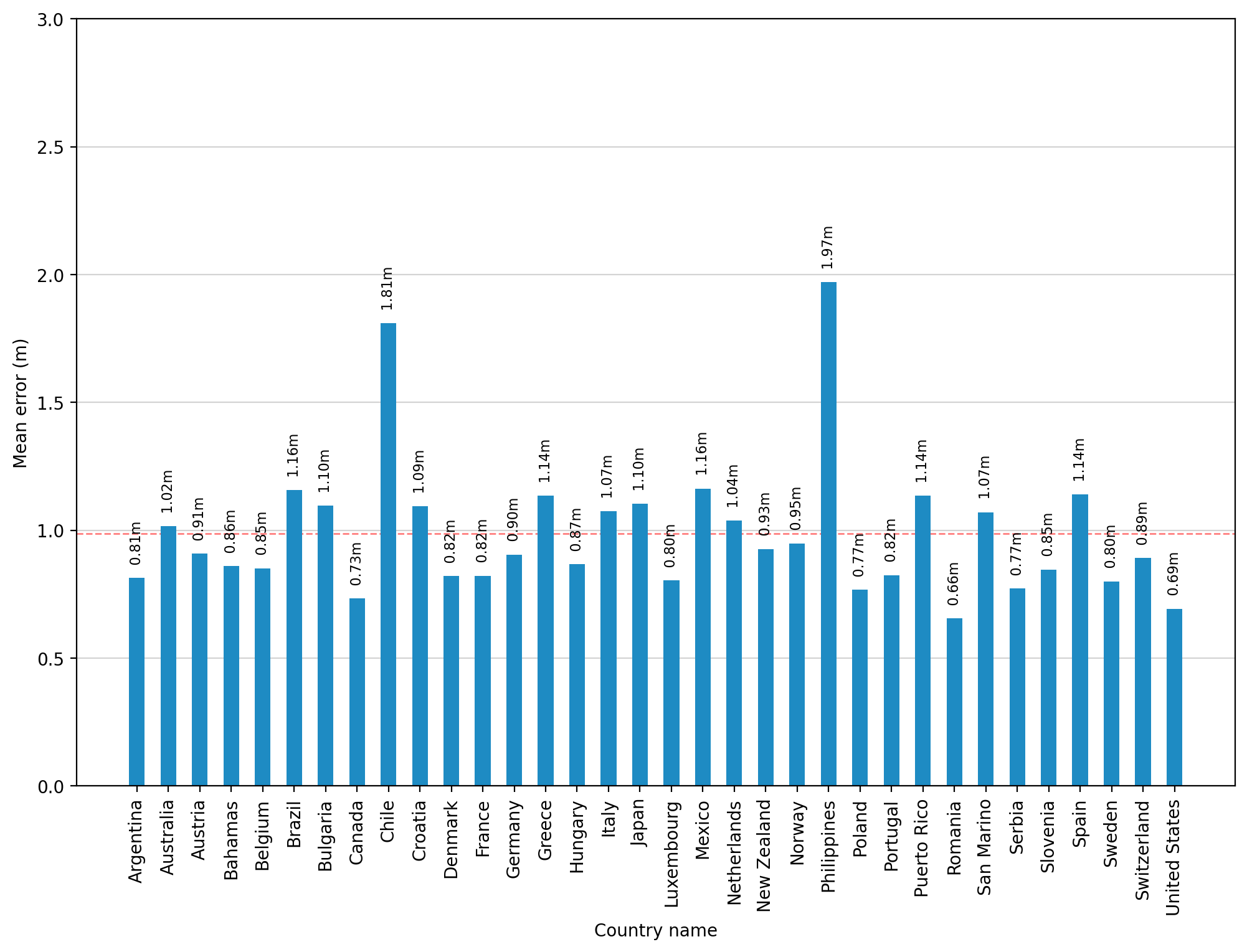}
  \caption{Per-country (RGB-only) height error distribution}
  \label{fig:country_mae_distrib}

  \centering
  \includegraphics[width=\textwidth]{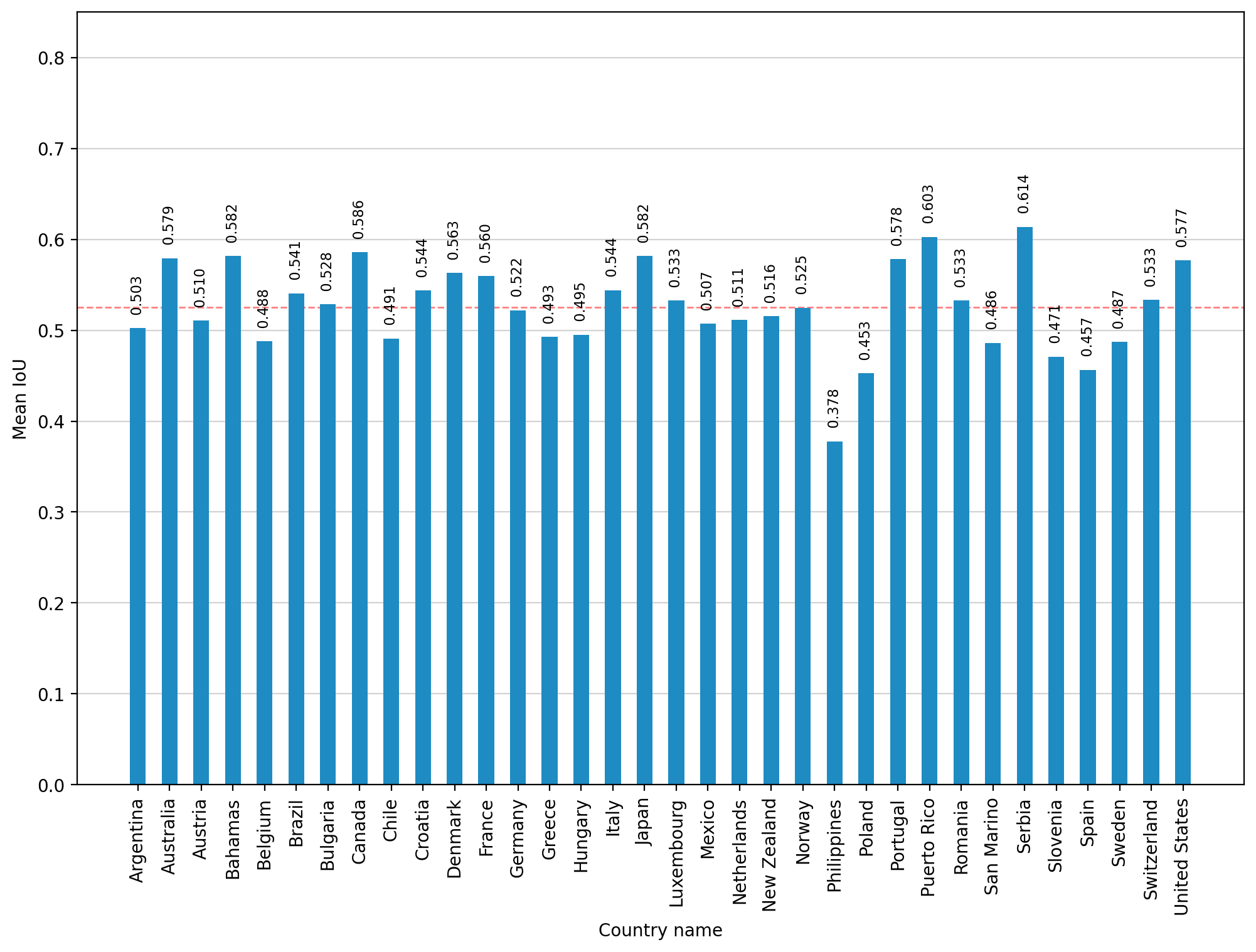}
  \caption{Per-country (RGB-only) roof segmentation IoU distribution}
  \label{fig:country_iou_distrib}
\end{figure}

Figures \ref{fig:country_mae_distrib} and \ref{fig:country_iou_distrib} show a per-country breakdown of the height error and roof segmentation performance for our RGB-only dataset (chosen as it has the greatest number of countries). We observe that the error variation between countries is small (with the exception of Chile and the Philippines, for which our ground-truth aerial imagery is atypically noisy - likely explaining their outlier status) and conclude that the model is able to adapt well to different regions and styles of housing.

\section{Qualitative results}

\begin{figure}[h!]
    \centering
    \begin{tabular}{ccc}
        \subcaptionbox{GT Aerial RGB\label{fig:subfig1}}{\includegraphics[width=0.3\textwidth]{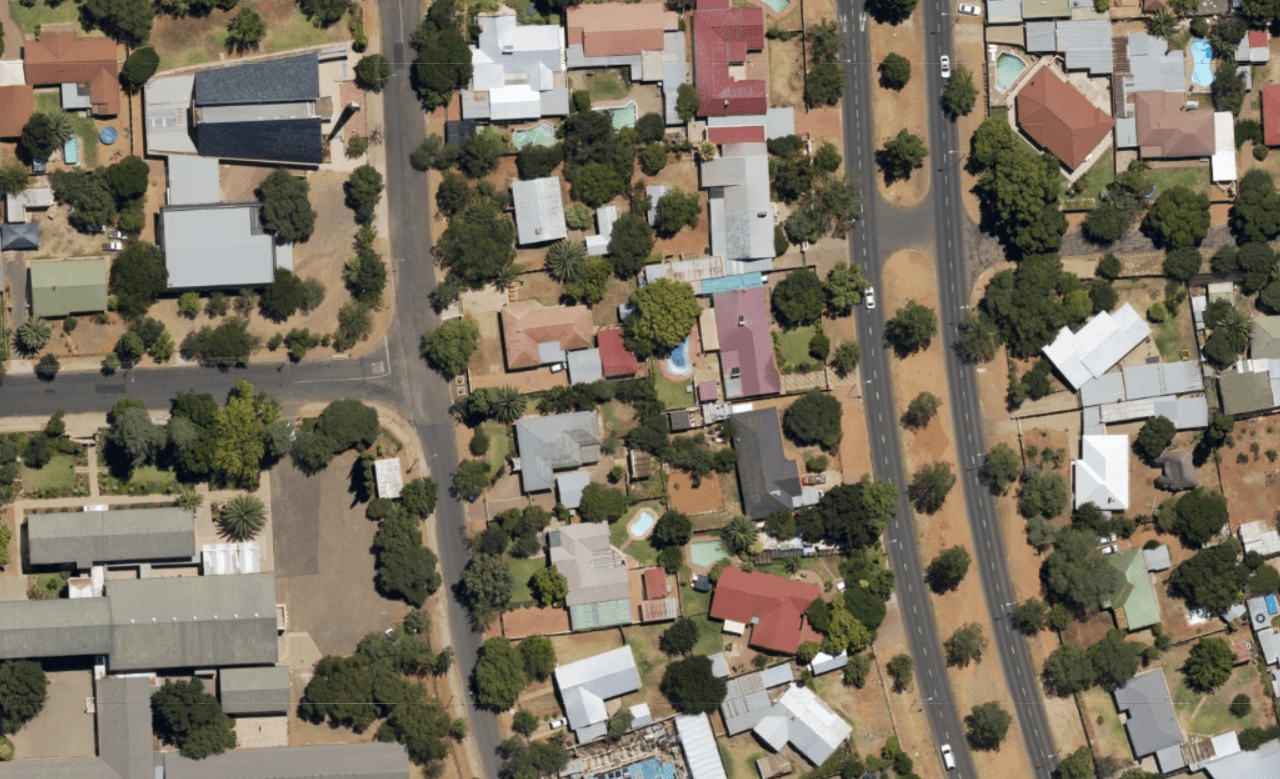}} &
        \subcaptionbox{GT Aerial DSM\label{fig:subfig2}}{\includegraphics[width=0.3\textwidth]{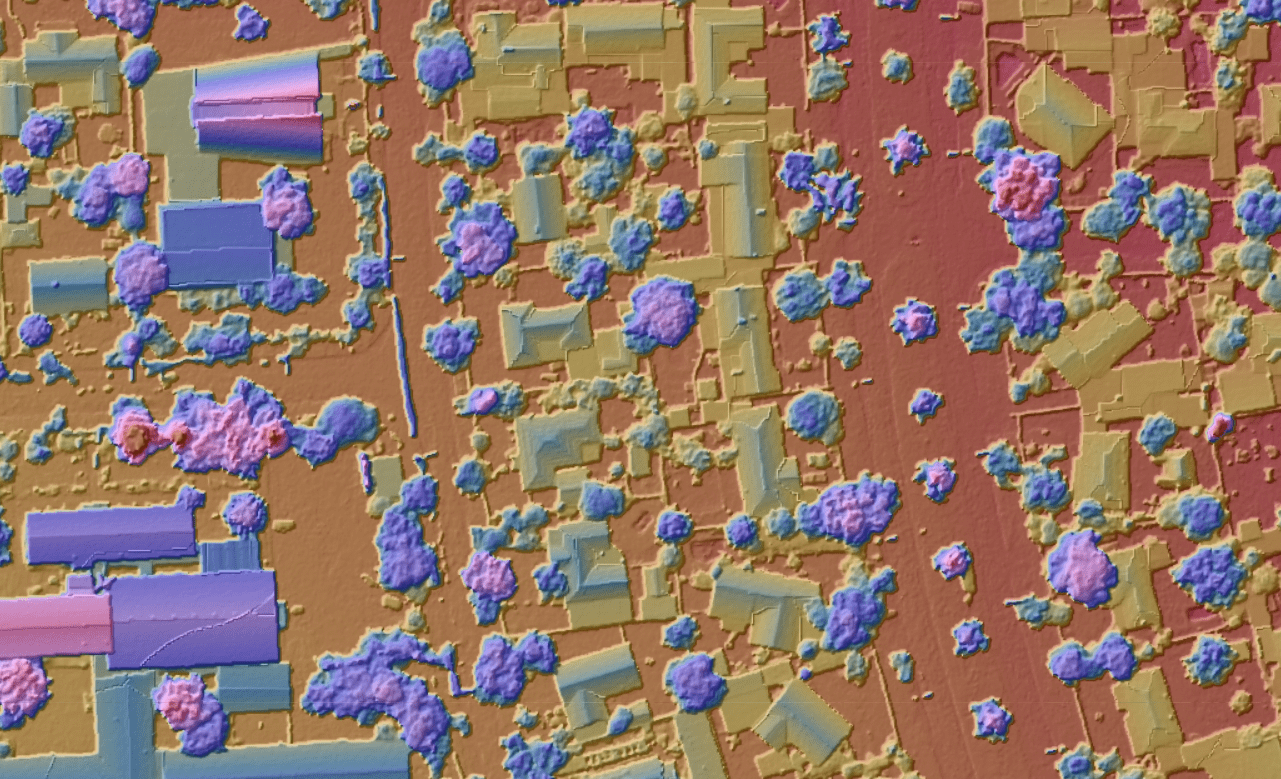}} &
        \subcaptionbox{GT Aerial roof segments\label{fig:subfig3}}{\includegraphics[width=0.3\textwidth]{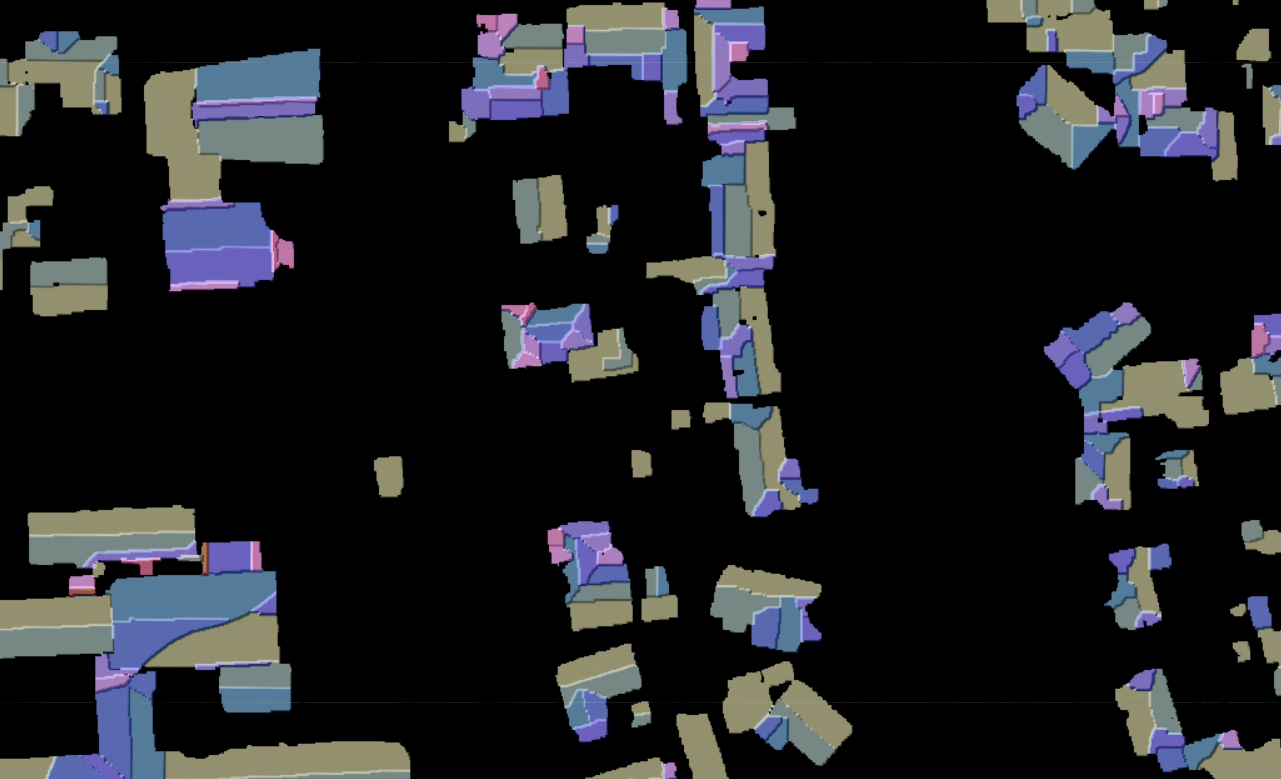}} \\\\

        \subcaptionbox{Satellite RGB\label{fig:subfig4}}{\includegraphics[width=0.3\textwidth]{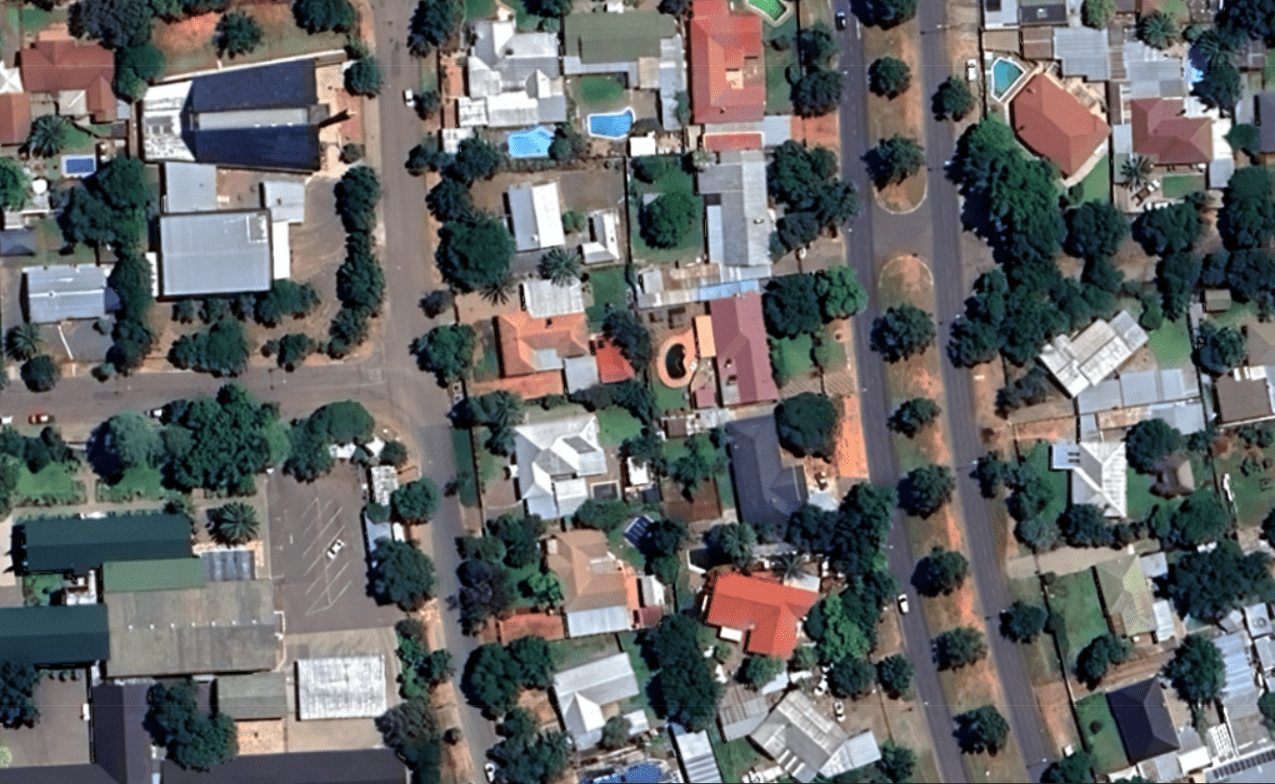}} &
        \subcaptionbox{Model output DSM\label{fig:subfig5}}{\includegraphics[width=0.3\textwidth]{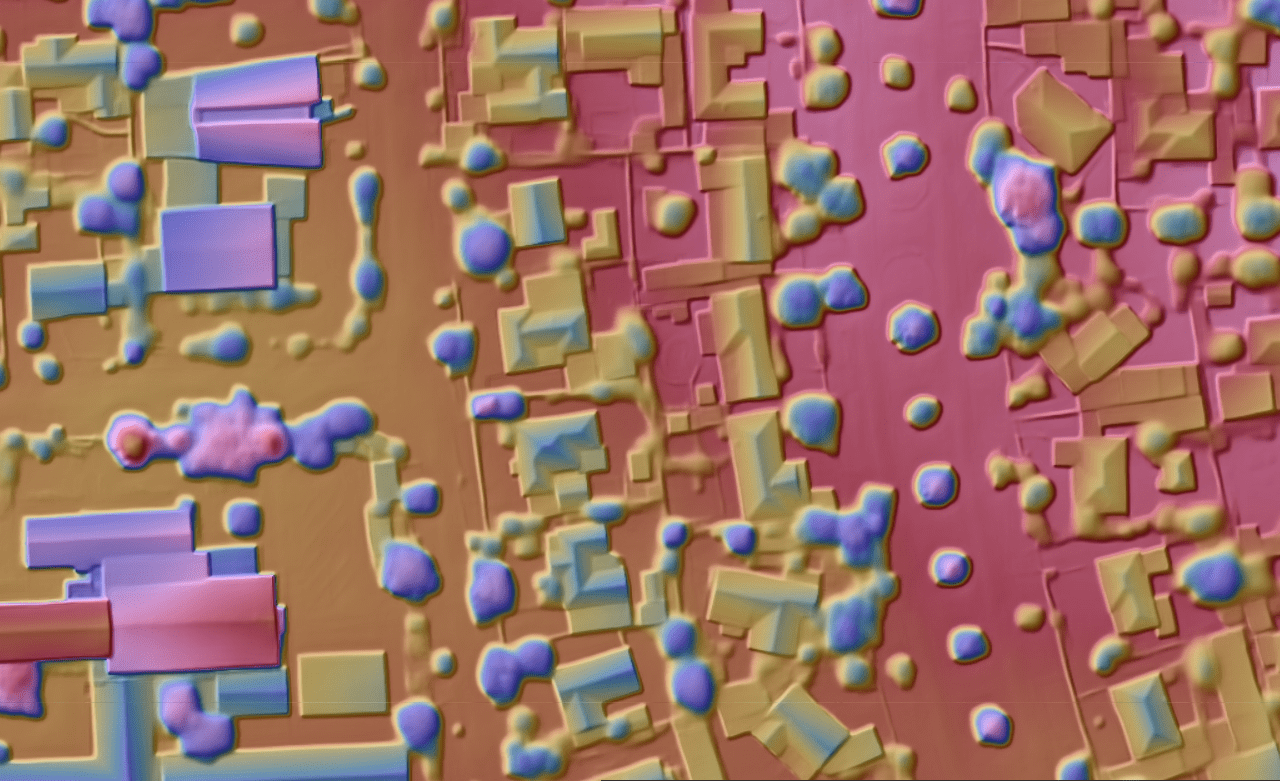}} &
        \subcaptionbox{Model output roof segments\label{fig:subfig6}}{\includegraphics[width=0.3\textwidth]{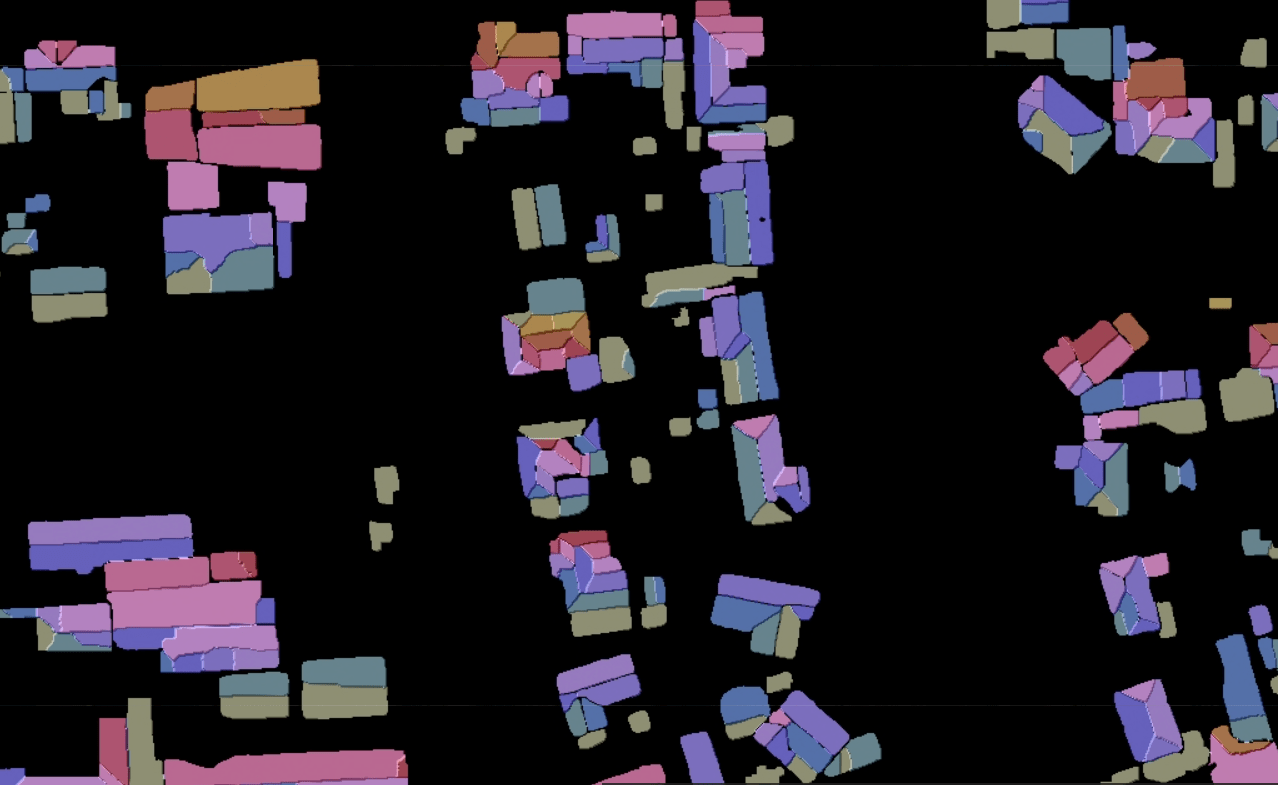}}
    \end{tabular}
    \caption{Comparison of sample model outputs with ground truth (GT) data. Location: Bloemfontein, South Africa}
    \label{fig:aerial_satellite_comparison}
\end{figure}

\begin{figure}[h!]
    \centering

    % Column captions at the top
    \begin{minipage}{0.24\textwidth}
        \centering
        \textbf{Off-nadir input RGB}
    \end{minipage}%
    \hfill
    \begin{minipage}{0.24\textwidth}
        \centering
        \textbf{Output nadir RGB}
    \end{minipage}%
    \hfill
    \begin{minipage}{0.24\textwidth}
        \centering
        \textbf{Output nadir DSM}
    \end{minipage}%
    \hfill
    \begin{minipage}{0.24\textwidth}
        \centering
        \textbf{Output nadir roof segments}
    \end{minipage}
    
    \vspace{0.1cm} % Space between column captions and first row
    
    % Row 1
    \begin{minipage}{0.24\textwidth}
        \centering
        \includegraphics[width=\textwidth]{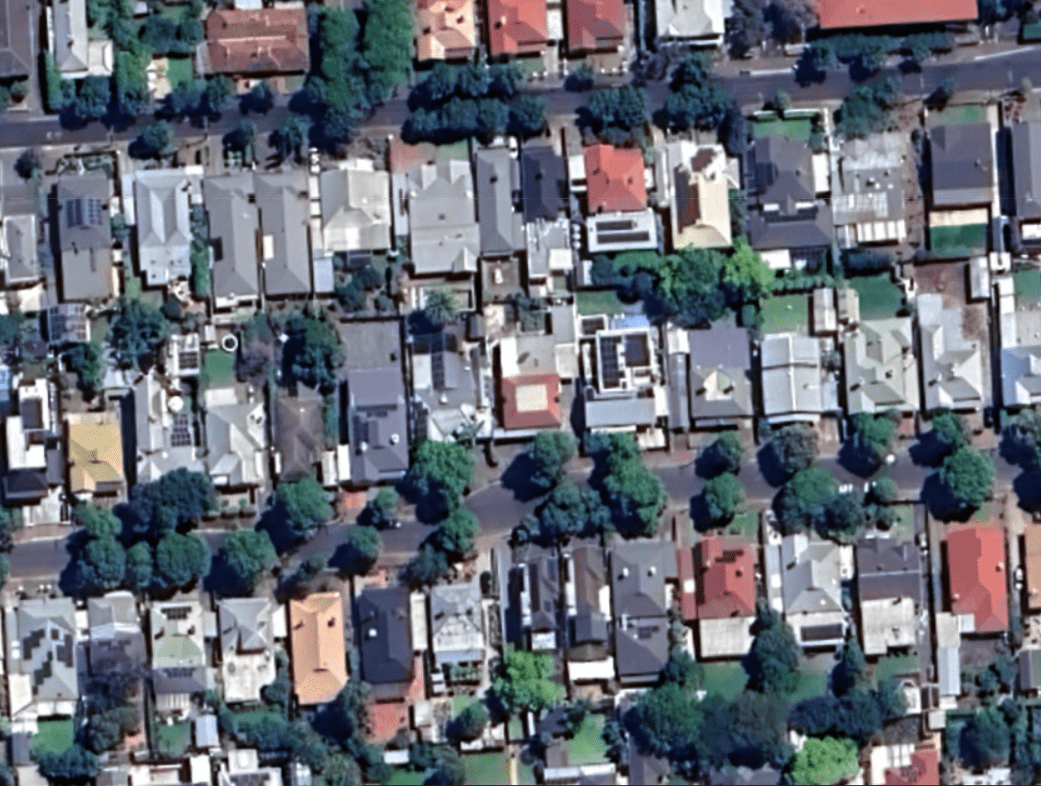}
    \end{minipage}%
    \hfill
    \begin{minipage}{0.24\textwidth}
        \centering
        \includegraphics[width=\textwidth]{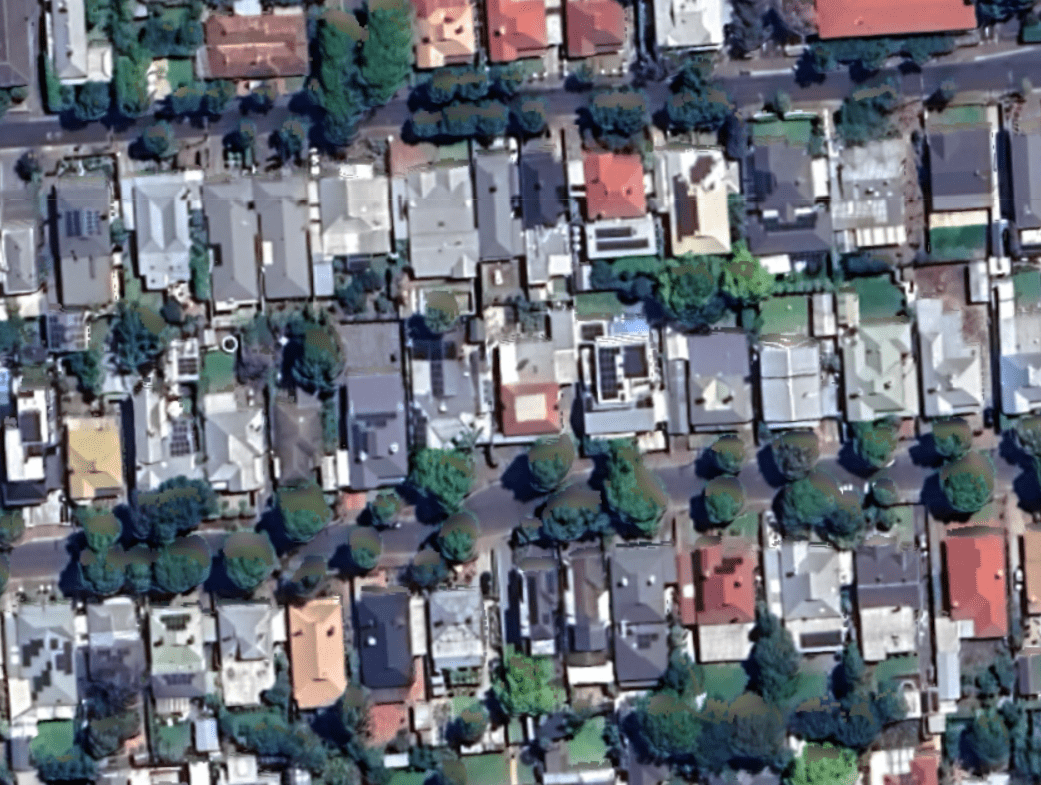}
    \end{minipage}%
    \hfill
    \begin{minipage}{0.24\textwidth}
        \centering
        \includegraphics[width=\textwidth]{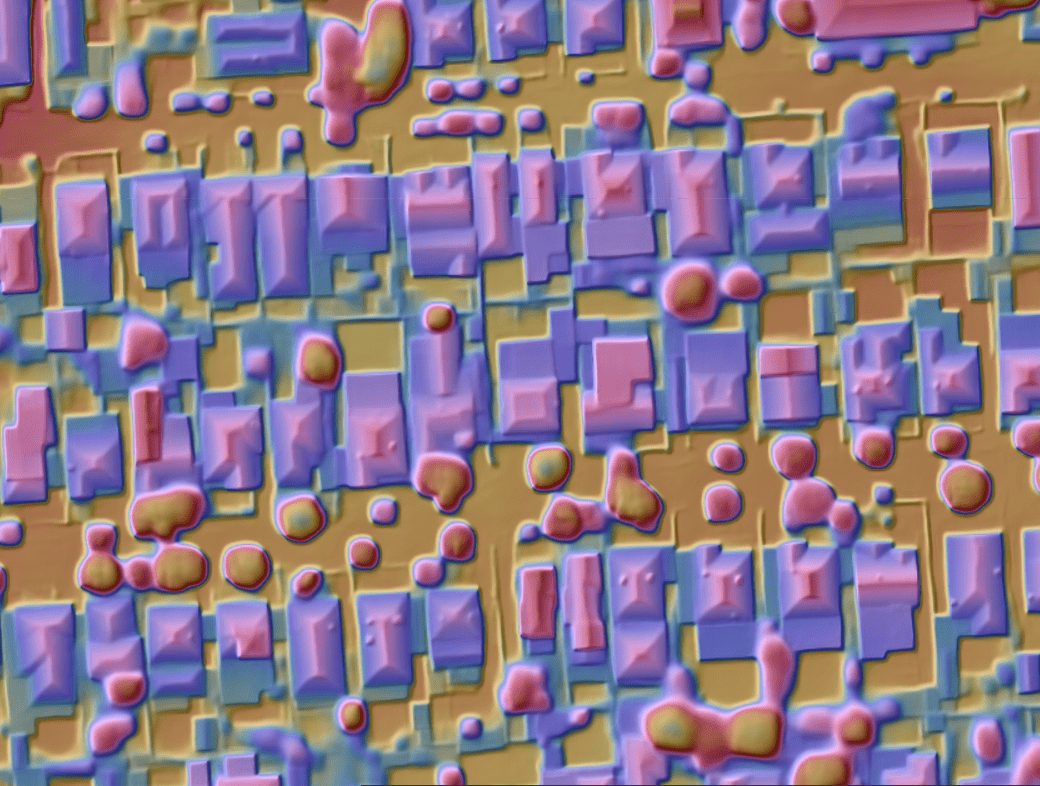}
    \end{minipage}%
    \hfill
    \begin{minipage}{0.24\textwidth}
        \centering
        \includegraphics[width=\textwidth]{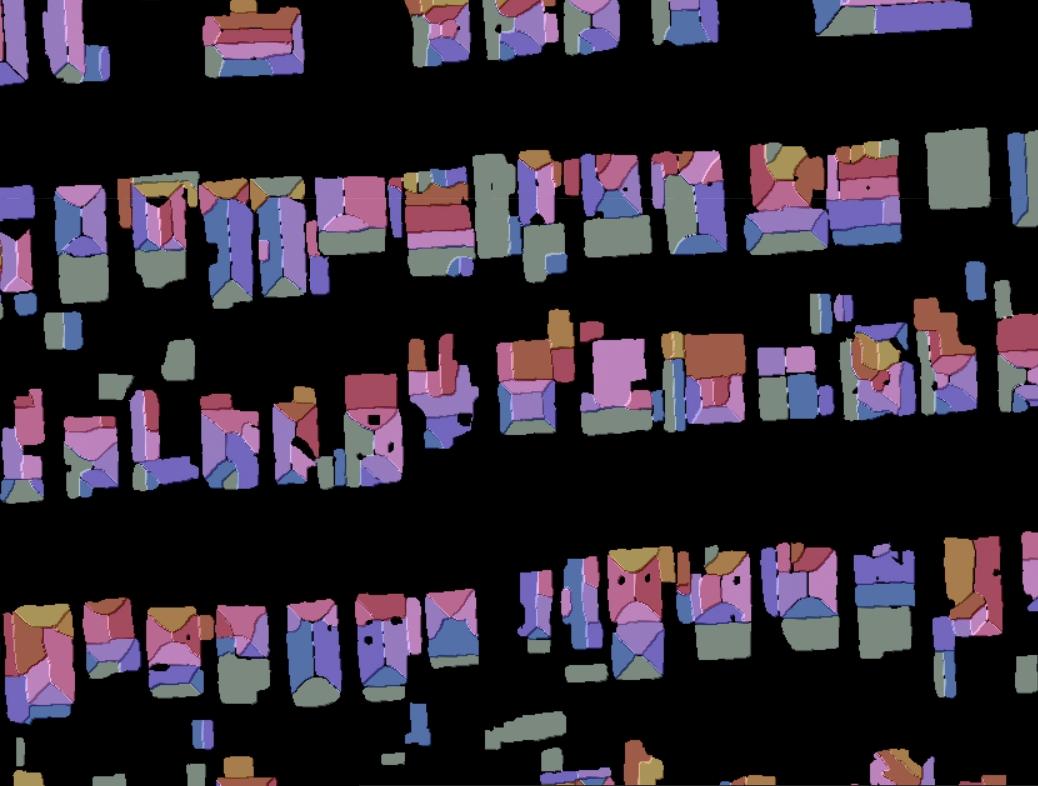}
    \end{minipage}

    \vspace{0.1cm} % Space between rows
    \centering
    Location: Adelaide, Australia.

    \vspace{0.1cm} % Space between rows

    % Row 2
    \begin{minipage}{0.24\textwidth}
        \centering
        \includegraphics[width=\textwidth]{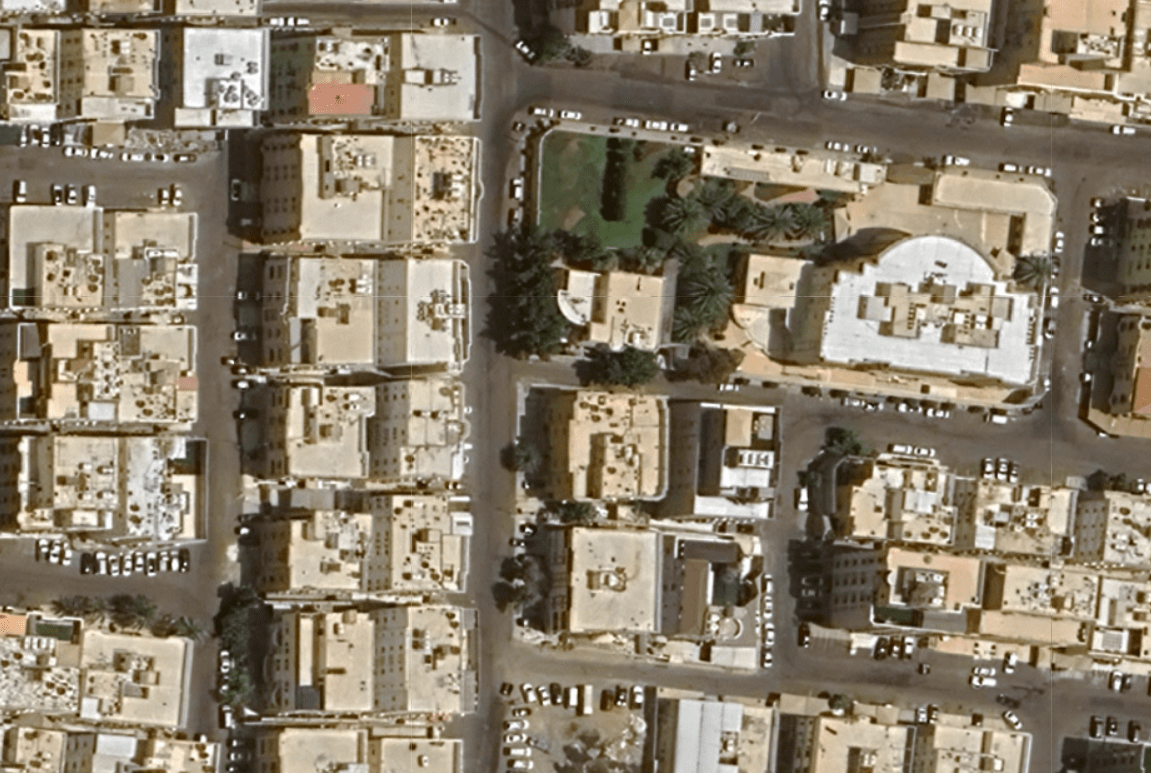}
    \end{minipage}%
    \hfill
    \begin{minipage}{0.24\textwidth}
        \centering
        \includegraphics[width=\textwidth]{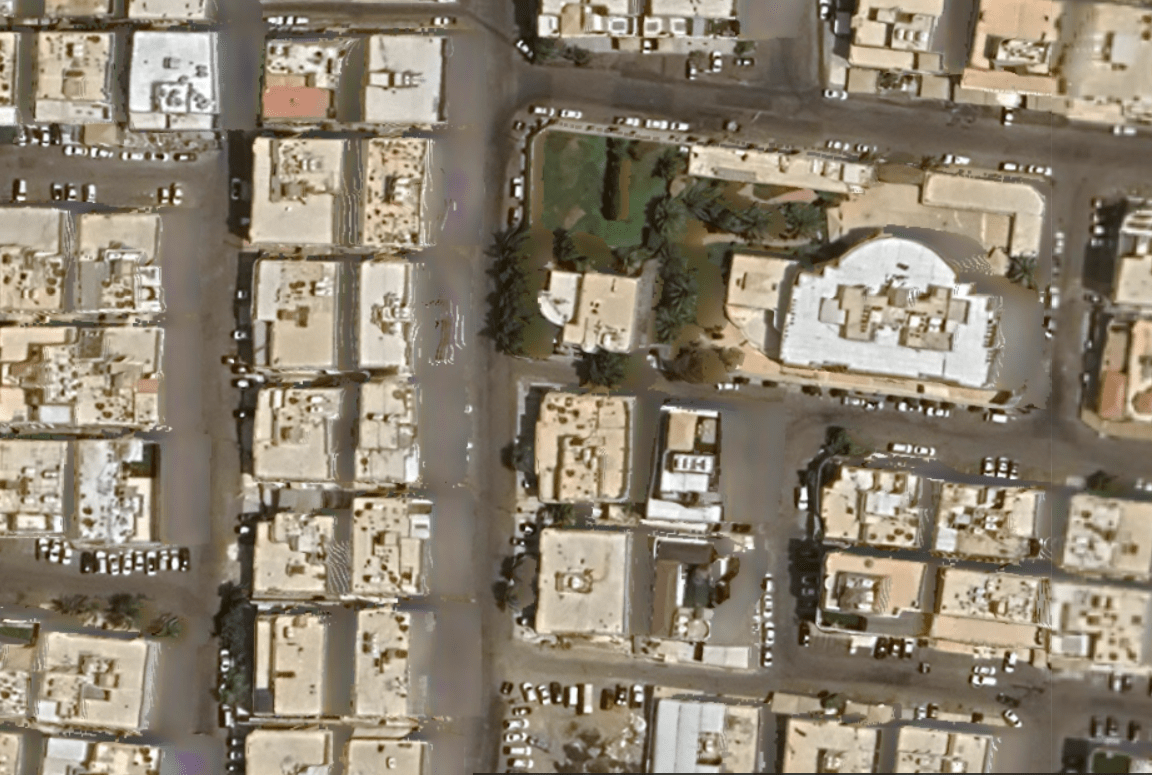}
    \end{minipage}%
    \hfill
    \begin{minipage}{0.24\textwidth}
        \centering
        \includegraphics[width=\textwidth]{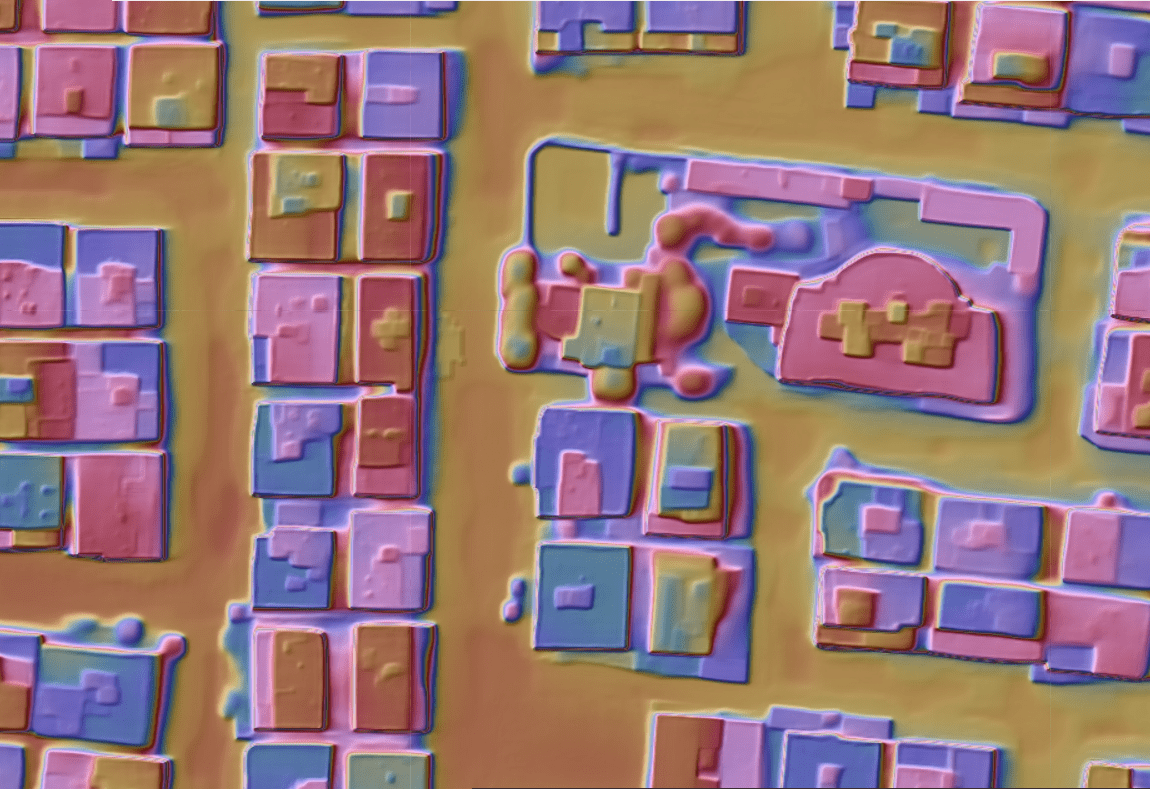}
    \end{minipage}%
    \hfill
    \begin{minipage}{0.24\textwidth}
        \centering
        \includegraphics[width=\textwidth]{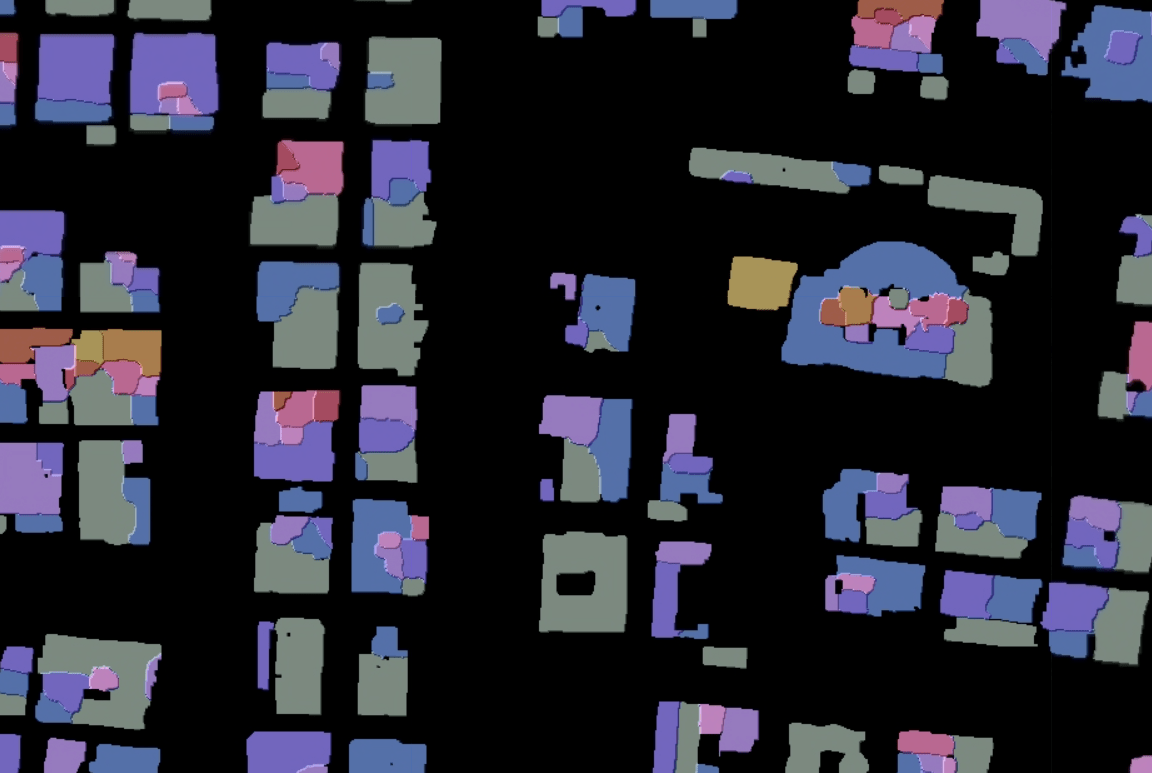}
    \end{minipage}

    \vspace{0.1cm} % Space between rows
    \centering
    Location: Jeddah, Saudi Arabia.

    \vspace{0.1cm} % Space between rows

    % Row 3
    \begin{minipage}{0.24\textwidth}
        \centering
        \includegraphics[width=\textwidth]{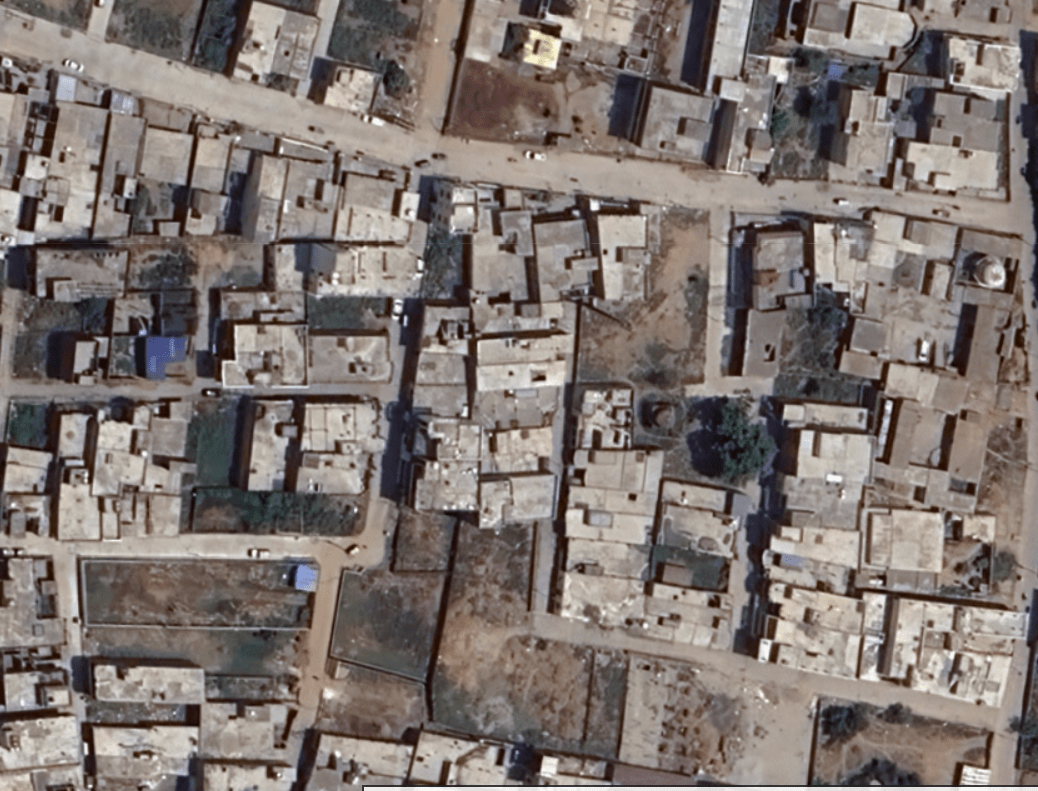}
    \end{minipage}%
    \hfill
    \begin{minipage}{0.24\textwidth}
        \centering
        \includegraphics[width=\textwidth]{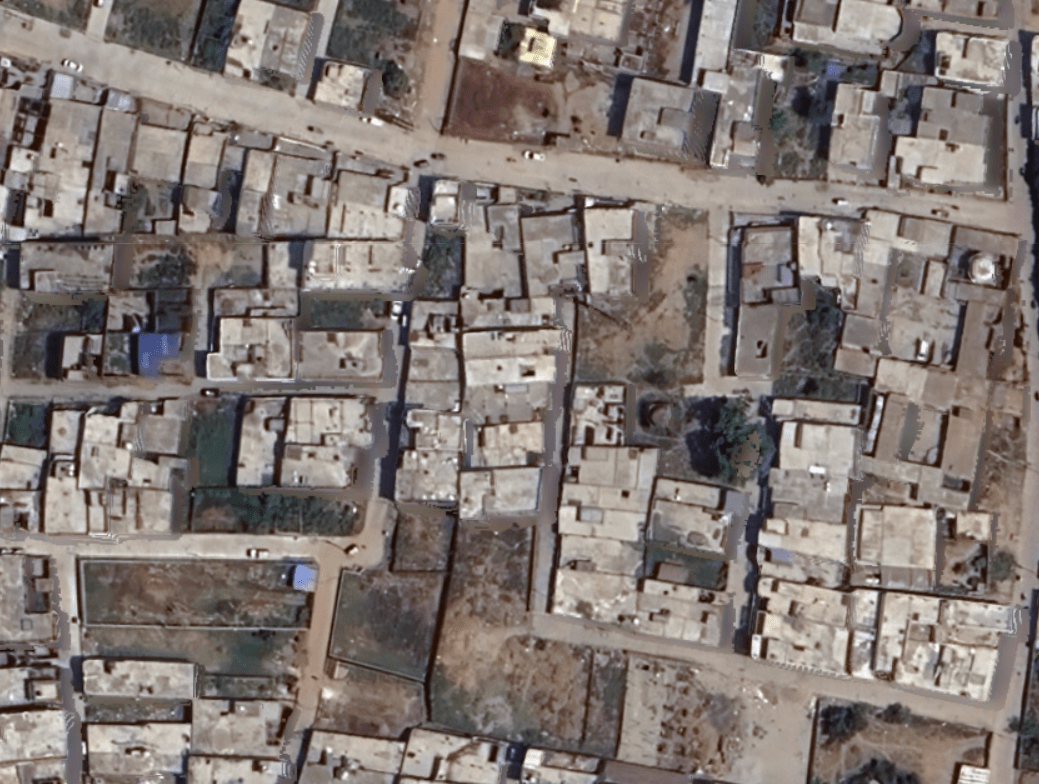}
    \end{minipage}%
    \hfill
    \begin{minipage}{0.24\textwidth}
        \centering
        \includegraphics[width=\textwidth]{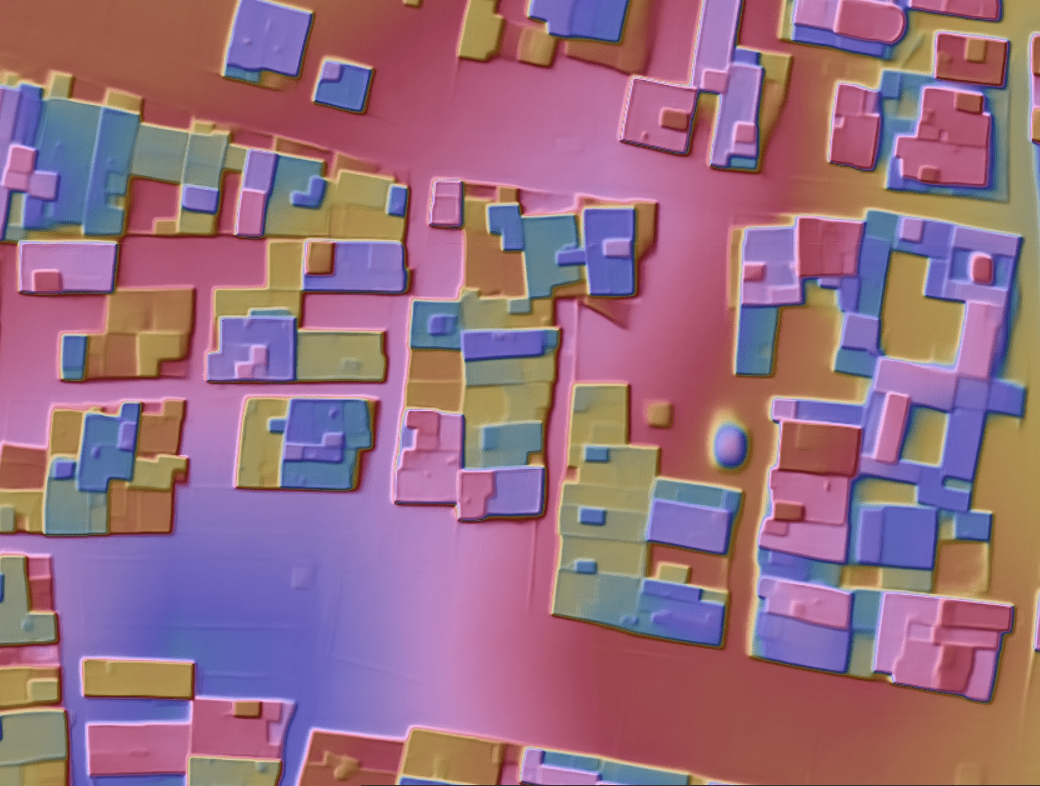}
    \end{minipage}%
    \hfill
    \begin{minipage}{0.24\textwidth}
        \centering
        \includegraphics[width=\textwidth]{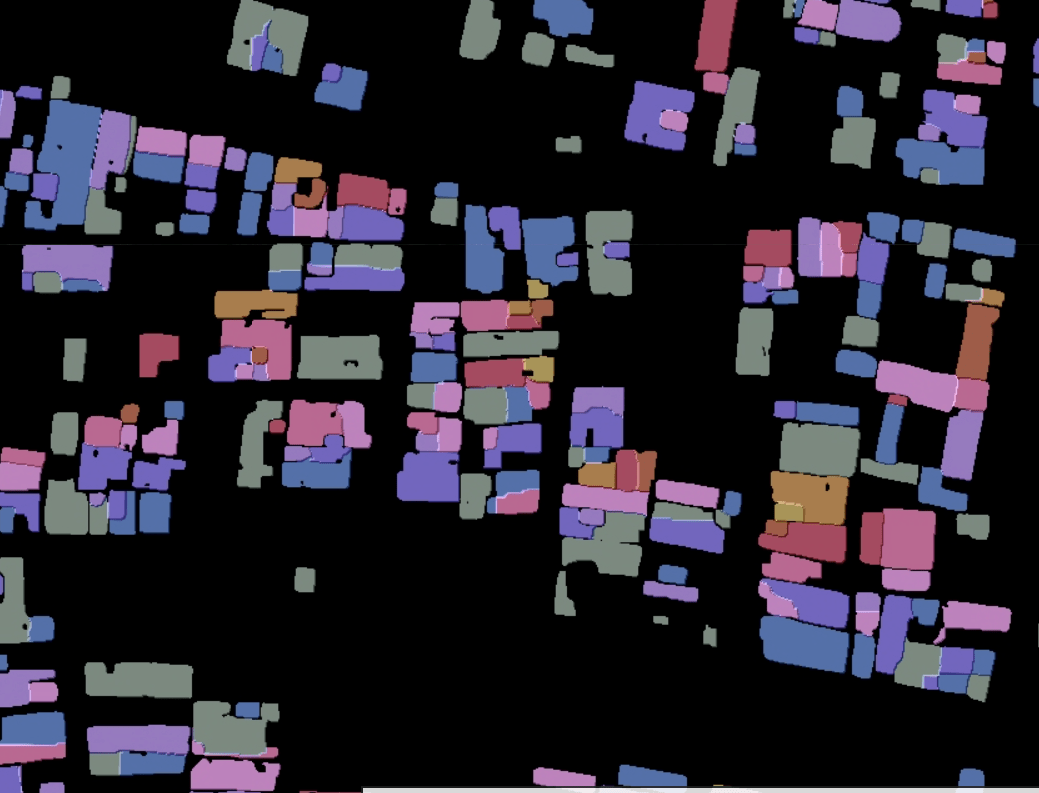}
    \end{minipage}

    \vspace{0.1cm} % Space between rows
    \centering
    Location: Ayodhya, India.

    \vspace{0.1cm} % Space between rows

    % Row 4
    \begin{minipage}{0.24\textwidth}
        \centering
        \includegraphics[width=\textwidth]{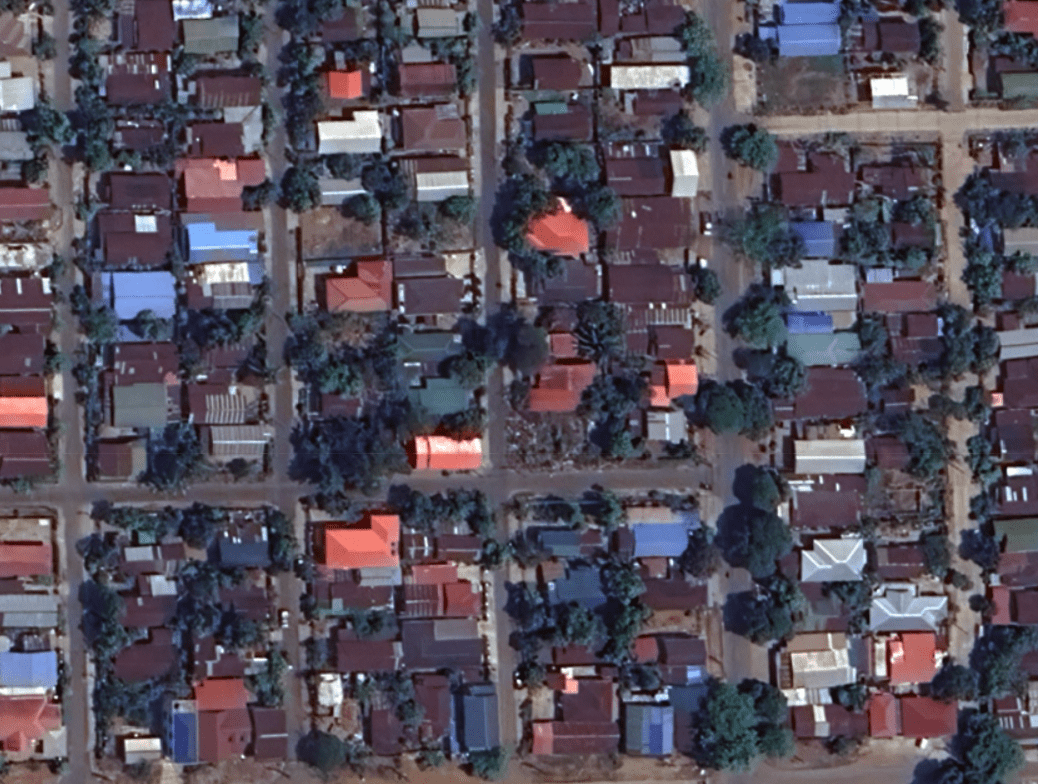}
    \end{minipage}%
    \hfill
    \begin{minipage}{0.24\textwidth}
        \centering
        \includegraphics[width=\textwidth]{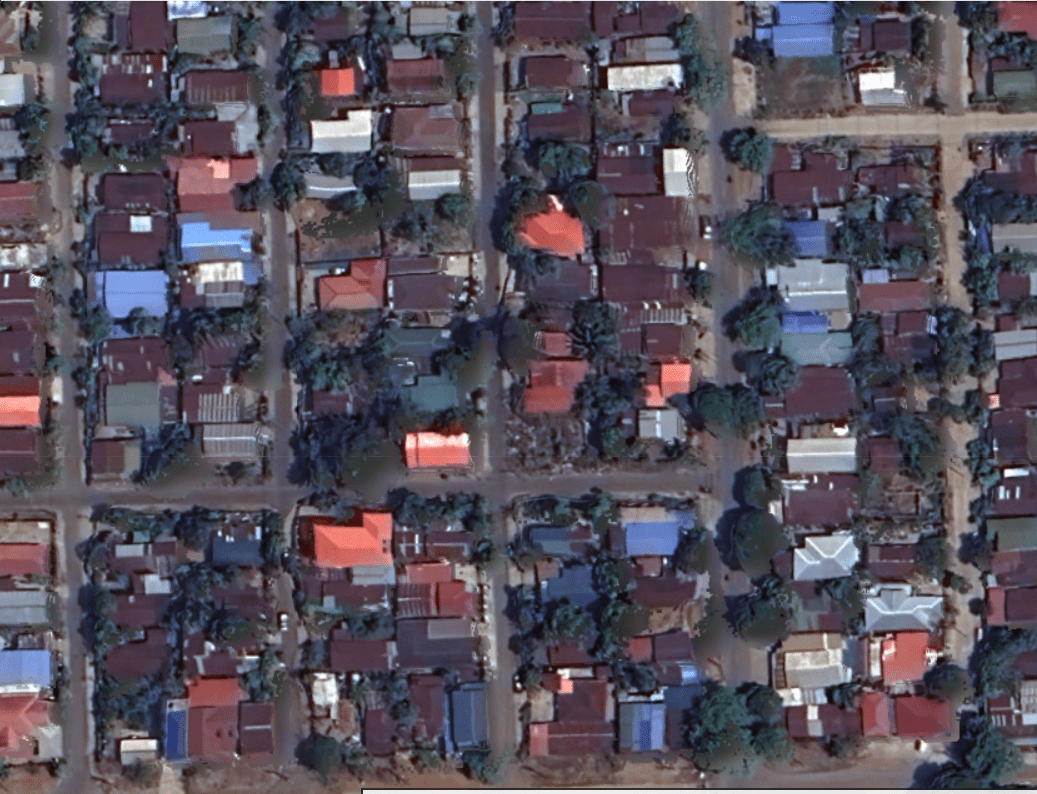}
    \end{minipage}%
    \hfill
    \begin{minipage}{0.24\textwidth}
        \centering
        \includegraphics[width=\textwidth]{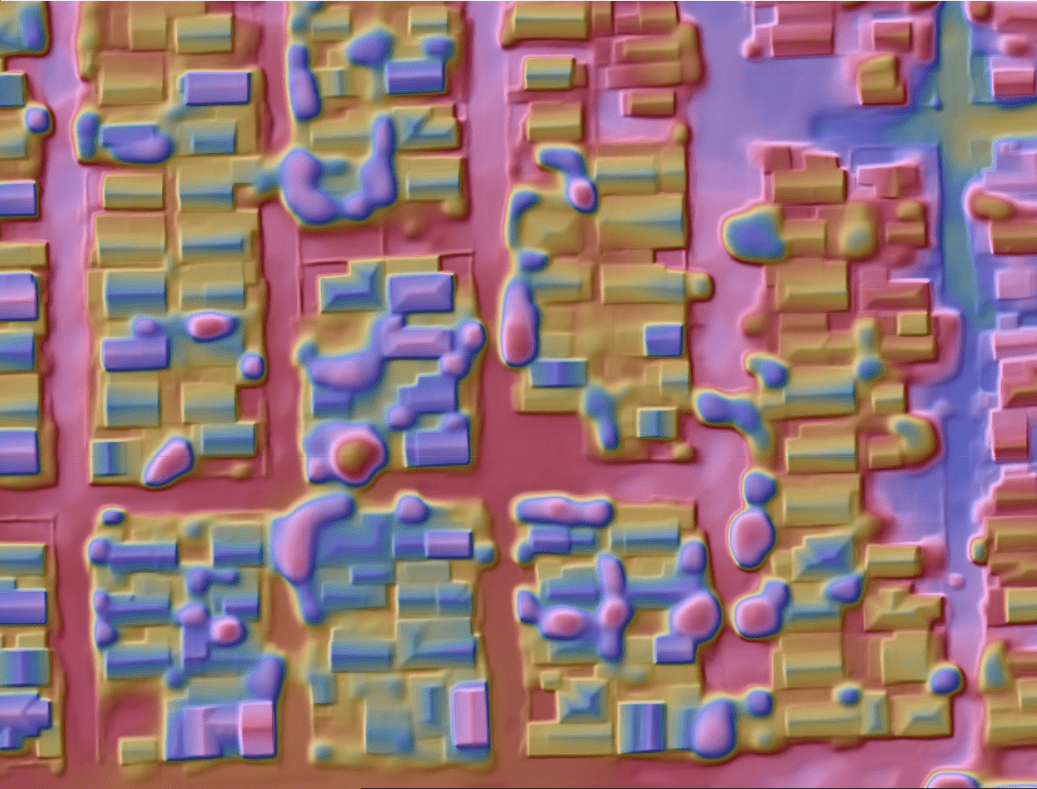}
    \end{minipage}%
    \hfill
    \begin{minipage}{0.24\textwidth}
        \centering
        \includegraphics[width=\textwidth]{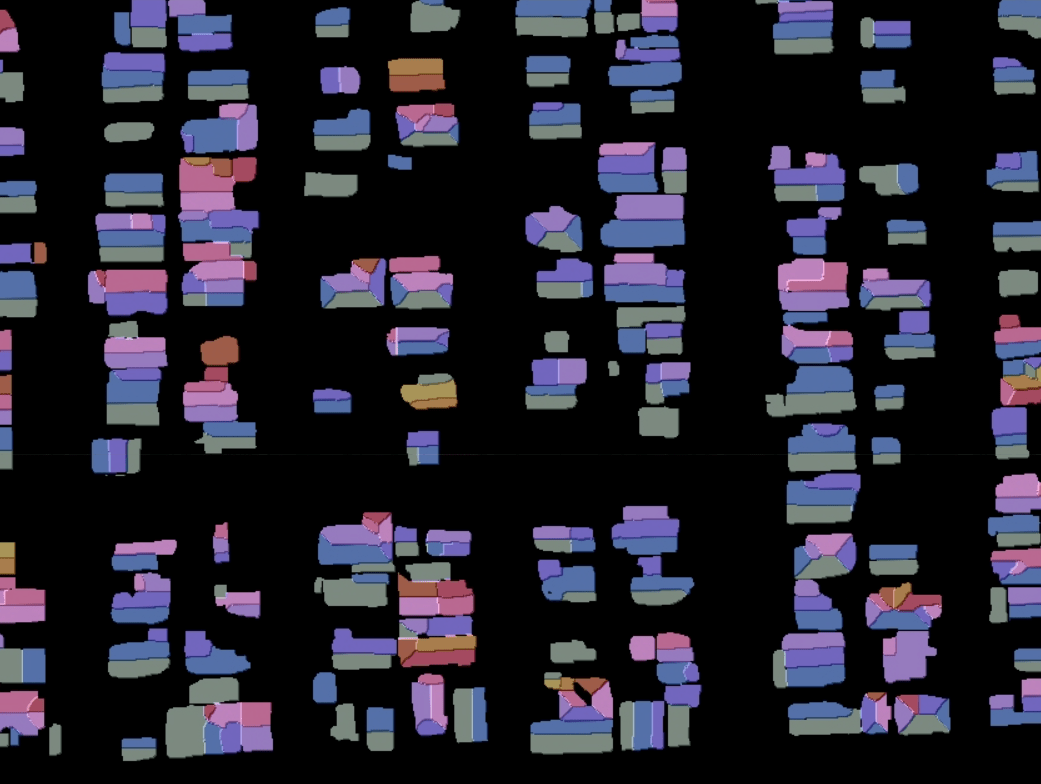}
    \end{minipage}

    \vspace{0.1cm} % Space between rows
    \centering
    Location: Mawlamyine, Myanmar.

    \vspace{0.1cm} % Space between rows

    % Row 5
    \begin{minipage}{0.24\textwidth}
        \centering
        \includegraphics[width=\textwidth]{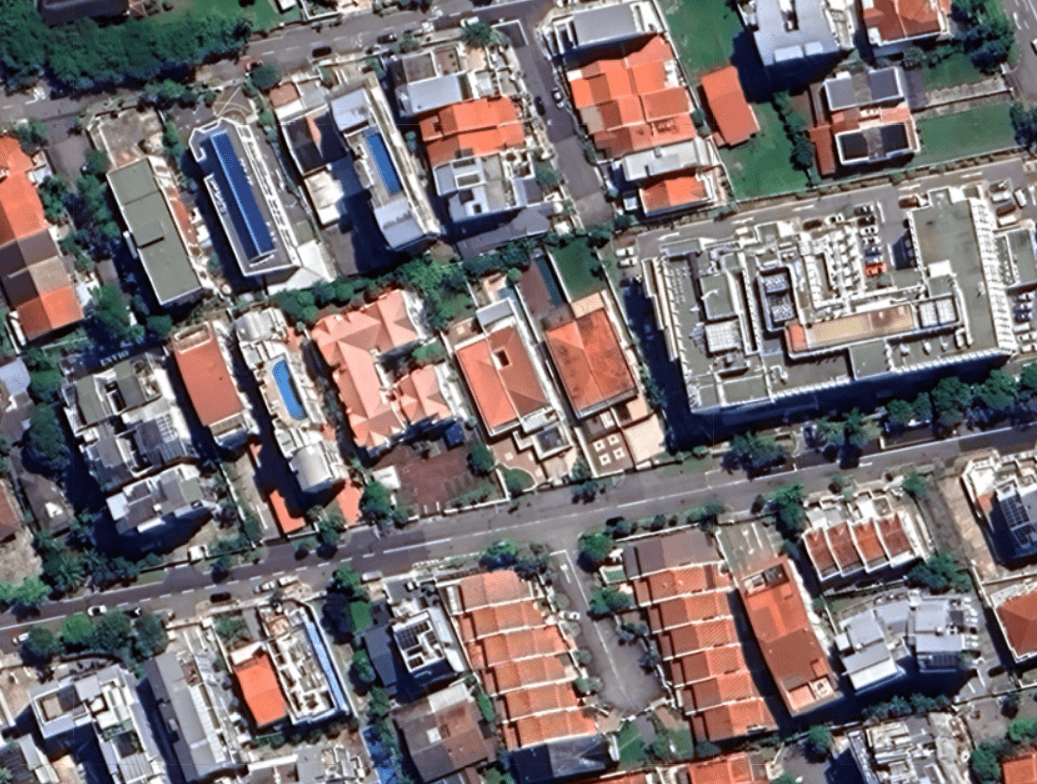}
    \end{minipage}%
    \hfill
    \begin{minipage}{0.24\textwidth}
        \centering
        \includegraphics[width=\textwidth]{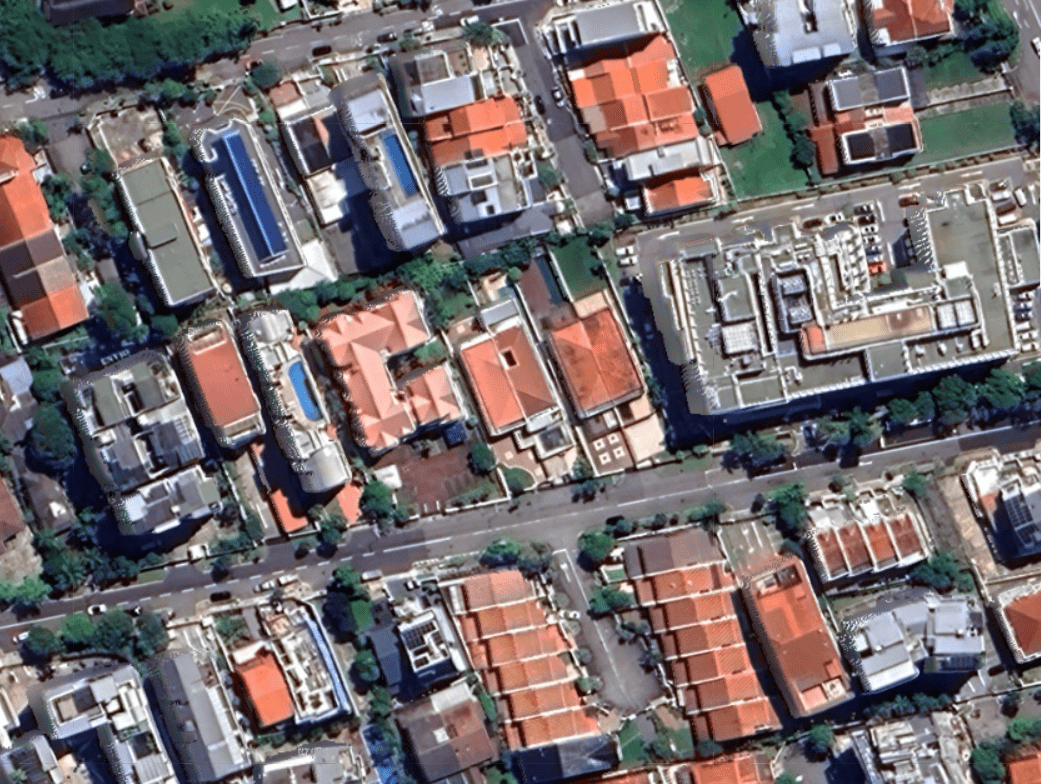}
    \end{minipage}%
    \hfill
    \begin{minipage}{0.24\textwidth}
        \centering
        \includegraphics[width=\textwidth]{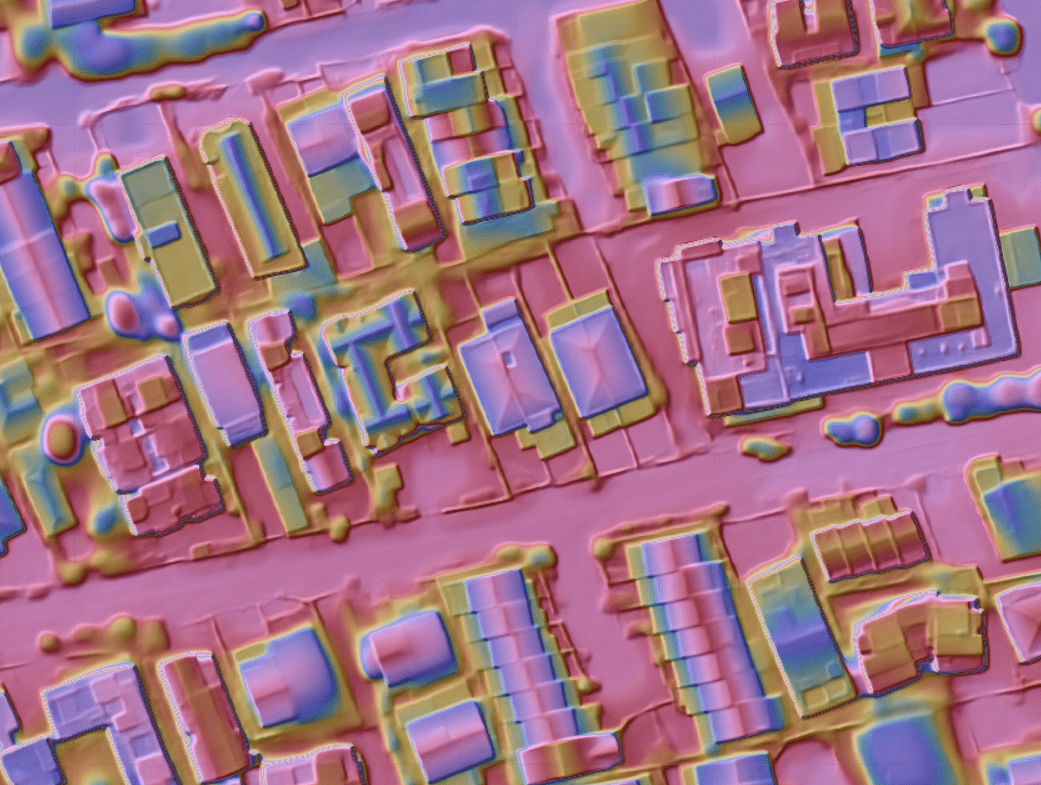}
    \end{minipage}%
    \hfill
    \begin{minipage}{0.24\textwidth}
        \centering
        \includegraphics[width=\textwidth]{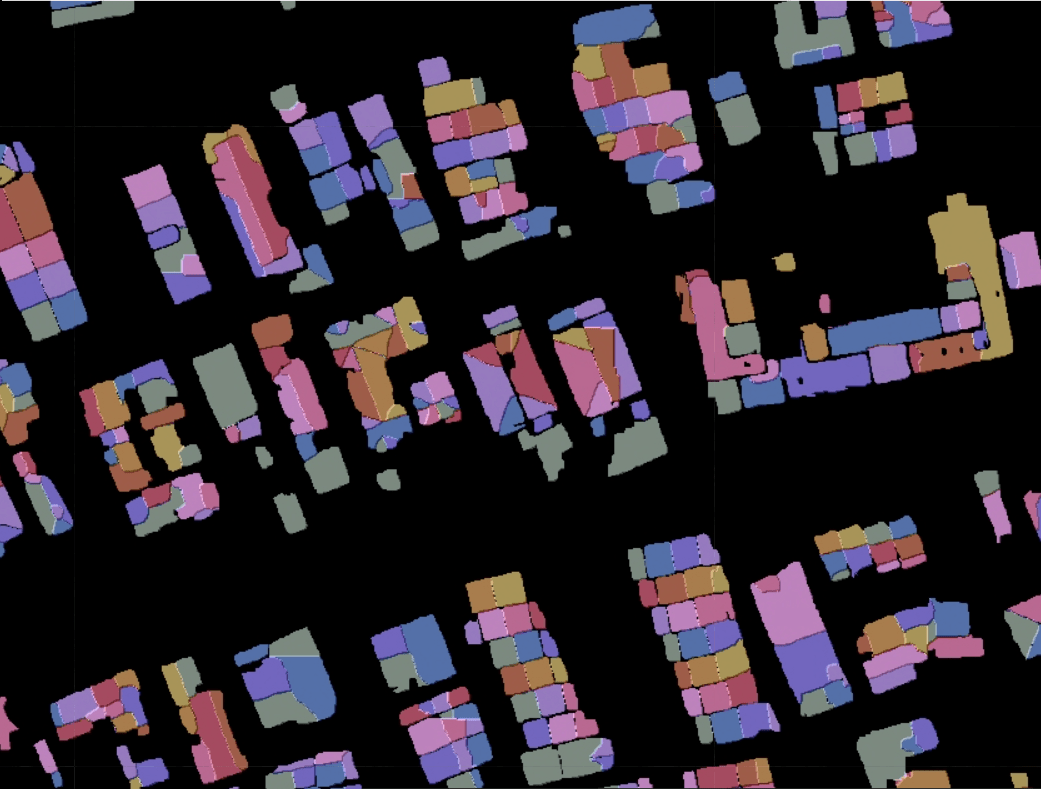}
    \end{minipage}

    \vspace{0.1cm} % Space between rows
    \centering
    Location: Singapore.

    \caption{Qualitative visualizations of model outputs from various geographies.}
    \label{fig:qualitative_results_vis}
\end{figure}

Figures \ref{fig:aerial_satellite_comparison} and \ref{fig:qualitative_results_vis} showcase sample model results. Figure \ref{fig:aerial_satellite_comparison} compares ground truth nadir aerial data with model predictions from off-nadir satellite imagery, highlighting accurate DSM and roof segment inference. Figure \ref{fig:qualitative_results_vis} further demonstrates performance on diverse off-nadir inputs, showcasing predicted nadir RGB, DSM, and roof segments.

%%%%%%%%%%%%%%%%%%%%%%%%%%%%%%%%%%%%%%%%%%%%%%%%%%%%%%%%%%%%

\end{document}